\documentclass[pdflatex,sn-basic,iicol]{sn-jnl}
\usepackage{graphicx}%
\usepackage{multirow}%
\usepackage{amsmath,amssymb,amsfonts}%
\usepackage{amsthm}%
\usepackage{mathrsfs}%
\usepackage[title]{appendix}%
\usepackage{xcolor}%
\usepackage{textcomp}%
\usepackage{manyfoot}%
\usepackage{booktabs}%
\usepackage{algorithm}%
\usepackage{algorithmicx}%
\usepackage{algpseudocode}%
\usepackage{listings}%
\usepackage{array}
\usepackage{rotating}
\usepackage[capitalise]{cleveref}
\usepackage{amsmath,amsfonts,bm}

\def\eqref#1{equation~\ref{#1}}
\def\1{\bm{1}}

\DeclareMathAlphabet{\mathsfit}{\encodingdefault}{\sfdefault}{m}{sl}
\SetMathAlphabet{\mathsfit}{bold}{\encodingdefault}{\sfdefault}{bx}{n}

\def\gL{{\mathcal{L}}}

\usepackage{overpic}
\usepackage{multirow}
\usepackage{multicol}
\usepackage{booktabs}
\usepackage{colortbl}
\definecolor{url}{RGB}{0,73,147}
\definecolor{mypink}{HTML}{bc4749}
\definecolor{mygray}{gray}{.9}

\usepackage{array}
\usepackage[table]{xcolor}

\usepackage{pifont}    % For \ding command
\usepackage{bbding}      

\usepackage{caption}
\usepackage{enumitem}
\usepackage{multirow}

\makeatletter
\newcommand{\thickhline}{%
 \noalign {\ifnum 0=`}\fi \hrule height 1pt
 \futurelet \reserved@a \@xhline
}
\makeatother

\usepackage{pifont}
\usepackage{xspace}

\theoremstyle{thmstyleone}%
\theoremstyle{thmstyletwo}%

\theoremstyle{thmstylethree}%
\usepackage{balance}

\usepackage{etoolbox}
\makeatletter
\patchcmd{\@maketitle}{\removelastskip\vskip24pt}{\removelastskip\vskip15pt}{}{}
\patchcmd{\@maketitle}{\removelastskip\vskip24pt}{\removelastskip\vskip15pt}{}{}
\renewcommand\abstracthead{\@startsection{section}{1}{\z@}%
  {-6pt \@plus0ex \@minus0ex}{3pt}{\abstractheadfont}}

\begin{document}

\title[MLLM-GS: Multimodal Large Language Model Prior-Guided Fast Novel View Synthesis in Extreme Environments]{MLLM-GS: Multimodal Large Language Model Prior-Guided Fast Novel View Synthesis in Extreme Environments}


\author[1]{\fnm{Zhiliang} \sur{Wu}}%\email{zhiliang.wu@ntu.edu.sg}

\author[2]{\fnm{Xinye} \sur{Zheng}}%\email{xinye.zheng@outlook.com}

\author*[3]{\fnm{Kun} \sur{Li}}\email{kun.li@uaeu.ac.ae}

\author*[2]{\fnm{Fei} \sur{Wang}}\email{jiafei127@gmail.com}

\author[2]{\fnm{Hao} \sur{Guo}}%\email{sifan10077@gmail.com}

\author[4]{\fnm{Yanyan} \sur{Wei}}%\email{weiyy@hfut.edu.cn}

\author[5]{\fnm{Guanghui} \sur{Yue}}

\author[6]{\fnm{Beier} \sur{Zhu}}%\email{beier.zhu@ustc.edu.cn}

\author[7]{\fnm{Wei} \sur{Zhou}}

\author[6]{\fnm{Xun} \sur{Yang}}%\email{xyang21@ustc.edu.cn}

{\affil[1]{\small\orgname{Nanyang Technological University}, \orgaddress{\city{Singapore}, \country{Singapore}}}

{\affil[2]{\small\orgname{Institute of Artificial Intelligence, Hefei Comprehensive National Science Center}, \orgaddress{\city{Hefei}, \country{China}}}

{\affil[3]{\small\orgname{United Arab Emirates University}, \orgaddress{\city{Abu Dhabi}, \country{United Arab Emirates}}}

\affil[4]{\small\orgname{Hefei University of Technology}, \orgaddress{\city{Hefei}, \country{China}}}

\affil[5]{\small\orgname{Shenzhen University}, \orgaddress{\city{Shenzhen}, \country{China}}}

\affil[6]{\small\orgname{University of Science and Technology of China}, \orgaddress{\city{Hefei}, \country{China}}}

\affil[7]{\small\orgname{Cardiff University}, \orgaddress{\city{Cardiff}, \country{United Kingdom}}}


\abstract{Novel view synthesis (NVS) in degraded scenes aims to reconstruct 3D scenes from multi-view observations affected by various degradations, and to generate clear and photorealistic images at unseen viewpoints.
Despite recent advances, existing methods often rely on degradation-specific models, limiting their generalization across diverse extreme environments. 
Generative models provide rich visual priors, but their outputs often suffer from pixel misalignment and color deviations, which can propagate into the reconstructed 3D scene and compromise reconstruction quality.
Furthermore, the high computational and memory costs of 3D reconstruction hinder the practical application of NVS in extreme environments.
To this end, we propose MLLM-GS, a multimodal large language model (MLLM) prior-guided framework for fast NVS in extreme environments. 
MLLM-GS consists of two branches: MLLM prior-guided view enhancement and fast 3D scene reconstruction.
The former leverages the rich visual priors of a frozen MLLM to enhance views under different severe degradations and introduces a lightweight correction network to reduce pixel shifts and color distortions in MLLM outputs.
With supervision from only a few paired clean views, this branch produces reliable enhanced views for all degraded observations, providing pseudo-GT views for subsequent 3D reconstruction.
The latter improves 3DGS efficiency by optimizing rasterization, Gaussian memory layout, and parameter updates, reducing redundant computation, memory-access overhead, and optimization costs.
Extensive experiments on multiple degraded scenes demonstrate that MLLM-GS effectively handles various severe degradations and achieves high-quality NVS with significantly lower reconstruction time and memory consumption.
\emph{Project Page}: \href{https://wzlwlm13.github.io/MLLM-GS/}{https://MLLM-GS.github.io}.}




\maketitle

\section{Introduction}

Novel view synthesis (NVS)~\citep{yu2026dcrr,liu2026realx3d,li2025wild3a} aims to model a 3D scene from existing multi-view observations and generate photorealistic images at unseen camera viewpoints. 
It requires not only accurate scene geometry and appearance, but also spatial consistency and high visual quality across different views~\citep{11419639,10891520}. 
By enabling 3D scene modeling and free-viewpoint rendering, NVS supports a wide range of applications, including virtual reality~\citep{yu2024immersivenerf}, digital twins~\citep{plizzari2024outlook}, robotic perception~\citep{lei2025gaussnav}, autonomous driving~\citep{11314796}, and digital content creation~\citep{dai2025go}.
With the rapid advancement of immersive visual applications and 3D intelligence, NVS has become an important topic in computer vision and computer graphics~\citep{gao2026cosurfgs,yu2026dcrr}.

Early NVS methods synthesize novel views by image interpolation~\citep{chen2023view}, view reprojection~\citep{seitz1996view}, and explicit geometric reconstruction~\citep{debevec2023modeling}. 
Such pipelines often rely on accurate camera poses, cross-view correspondences, and scene geometry.
As a result, they can produce noticeable rendering errors under large viewpoint variations, complex occlusions, or inaccurate geometry estimation. 
With the development of deep learning, researchers began to learn continuous scene representations with neural networks, reducing the dependence on explicit geometric reconstruction. 
Among these approaches, Neural Radiance Fields (NeRF)~\citep{mildenhall2021nerf} implicitly models the mapping from spatial positions and viewing directions to color and density, and uses volume rendering to synthesize high-quality novel views.
NeRF significantly improves the representation capacity and rendering fidelity in complex scenes, and many follow-up methods~\citep{chen2023structnerf,liu2026i2,levy2023seathru,cui2024aleth} have been developed based on NeRF.
However, NeRF requires dense sampling along camera rays and a large number of network evaluations, resulting in long optimization times and making real-time rendering difficult.

\begin{figure}[t!]
\centering
\includegraphics[width=1.0\linewidth]{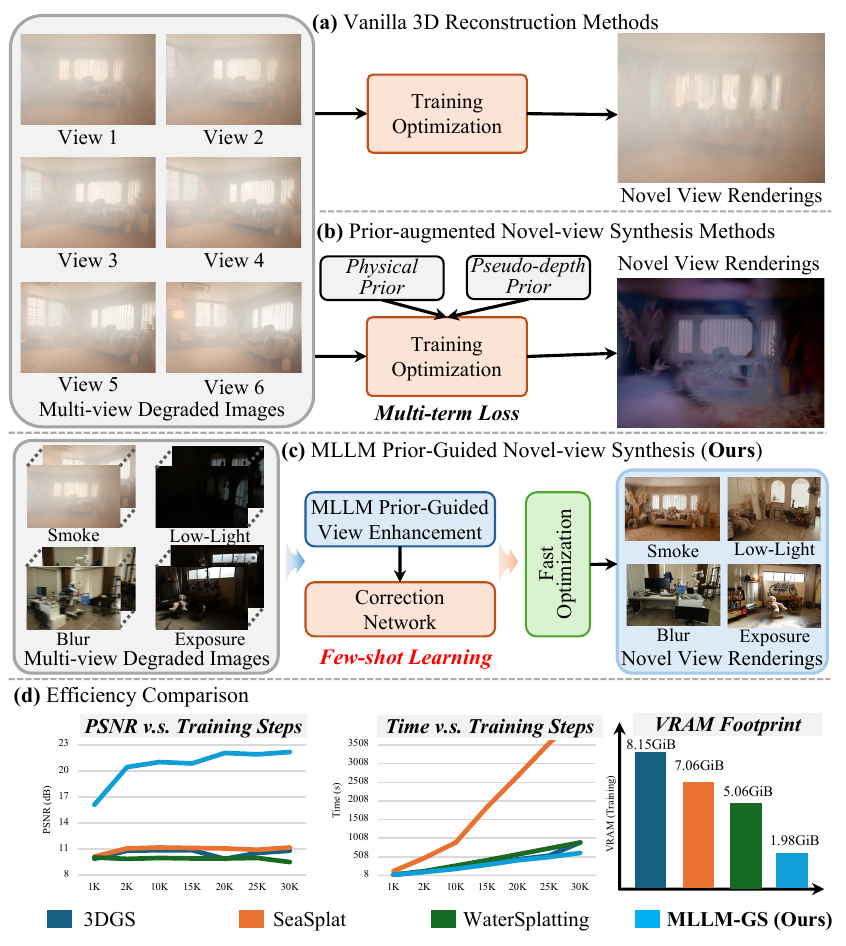}
\caption{Illustration of three different pipelines for NVS under degraded conditions. 
(a) Vanilla reconstruction, (b) prior‑augmented reconstruction, and (c) the proposed MLLM prior‑guided pipeline. 
(d) Efficiency comparison with previous representative methods. 
}
\label{fig:intro}
\vspace{-0.1cm}
\end{figure}

Recently, 3D Gaussian Splatting (3DGS) ~\citep{kerbl20233d} has emerged as an important approach for NVS. 
Unlike NeRF-based methods that represent a scene with implicit neural networks, 3DGS models scene geometry and appearance using a set of explicit 3D Gaussian primitives and directly projects them onto the image plane through differentiable rasterization.
Benefiting from its explicit scene representation and efficient splatting-based rendering, 3DGS significantly reduces optimization and rendering costs while maintaining high rendering quality.
Subsequent studies~\citep{yang2025seasplat,hahlbohm2026faster,ren2026fastgs,lee2026optimized} have further improved 3DGS in terms of rendering quality, representation capability, and computational efficiency, establishing it as a practical paradigm for high-quality and efficient NVS.

Despite these advances, existing 3DGS-based methods still heavily rely on high-quality multi-view observations~\citep{li2025watersplatting,cui2025luminance}.
In extreme environments involving smoke, low light, motion blur, and exposure variations, image structure, texture, and color information can be severely degraded. 
Such degradations weaken the visual features required for accurate geometry estimation and appearance modeling, leading to structural errors, missing textures, and color distortions in reconstructed scenes.
As illustrated in Fig.~\ref{fig:intro}, existing NVS methods~\citep{lin2025hqgs,liu2025deraings,lee2024deblurring,liu2025spatiotemporal} for degraded scenes mainly address these problems by introducing physical priors into the reconstruction process.
However, these methods typically rely on degradation-specific imaging models and assumptions. 
As different degradations follow distinct imaging mechanisms, such assumptions may not hold under unknown or mixed degradation conditions, limiting their generalization in extreme environments.
Consequently, existing approaches struggle to achieve robust NVS across diverse extreme environments within a unified framework.

An intuitive solution is to restore degraded images before performing 3D reconstruction.
However, conventional image restoration networks~\citep{yue2025text,wang2026neural,wu2026learning,long2026progressive} are typically trained on large-scale paired datasets and often specialize in particular degradation types.
Therefore, their performance can be limited by variations in degradation distributions or the lack of sufficient paired clean images.
This problem is particularly challenging in real-world extreme environments, where acquiring clean observations corresponding to all degraded viewpoints is often difficult. 
Recent generative models~\citep{11433514,li2026hybrid,zhu2025intelligent,jiang2026multi} offer a promising alternative by leveraging rich visual priors learned from large-scale data to recover plausible scene structures and textures under diverse degradation conditions. 
Nevertheless, their generation processes generally lack strict pixel-level constraints and may introduce spatial misalignment, altered textures, and color deviations. 
Although the restored images may appear visually realistic, pixel misalignment and color deviations can make them inconsistent with the underlying scene geometry and appearance~\citep{zhao2026mitigating,yang2025preliminary}.
When these images are used to supervise 3D reconstruction, such errors can propagate into the learned 3D representation, leading to structural distortions and appearance inconsistencies in novel-view renderings.
Therefore, effective NVS in extreme environments requires not only recovering visually plausible content but also improving the pixel-level fidelity of restored images under limited clean-image supervision.

Beyond reconstruction quality, computational efficiency remains another important challenge for NVS in severely degraded scenes. 
Although view enhancement improves the quality of degraded observations, residual artifacts and cross-view inconsistencies can still disrupt Gaussian optimization and induce redundant densification~\citep{lyu2026faster,zhou2026gradient}. 
The resulting increase in Gaussian primitives can further exacerbate the computational and memory overhead of standard 3DGS~\citep{kerbl20233d}.
First, its conservative screen-space bounding regions include many pixels receiving negligible Gaussian contributions, resulting in unnecessary rasterization operations.
Second, its adaptive densification and pruning can scatter spatially neighboring Gaussians across memory, reducing memory-access efficiency. 
Third, updating different Gaussian attributes through separate optimization operations introduces additional kernel launches and global memory accesses.
Although recent acceleration methods~\citep{ren2026fastgs,hahlbohm2026faster,hanson2025speedy} have improved 3DGS efficiency, these strategies have mainly been developed and evaluated using high-quality observations.
Their effectiveness in degraded scenes remains insufficiently explored.
Therefore, achieving both high-quality reconstruction and low computational cost under diverse severe degradations remains challenging.

In this paper, we propose MLLM-GS, a multimodal large language model (MLLM) prior-guided framework for fast NVS in extreme environments. 
MLLM-GS consists of two branches: MLLM prior-guided view enhancement and fast 3D scene reconstruction.
The former leverages the visual priors of MLLMs to generate reliable pseudo-GT views from degraded observations, thereby alleviating the scarcity of reliable supervision for 3D reconstruction.
Specifically, we employ a frozen MLLM to generate corresponding enhanced views from degraded observations under diverse degradation conditions.
In this way, the framework exploits the rich visual priors of the MLLM to handle multiple degradations in a unified manner, avoiding the need for degradation-specific restoration models.
Indeed, a similar prior-based enhancement strategy was also validated in the NTIRE 2026 3D Restoration and Reconstruction Challenge~\citep{liu2026ntire}, where our Smoke-GS~\citep{zheng20263d} solution achieved second place in Track 2.
However, the MLLM generation process lacks strict pixel-level constraints and may introduce pixel shifts and color distortions.
To address this issue, we further design a lightweight correction network. 
The network refines the MLLM-enhanced results in a color space that decouples luminance and chrominance, enabling more effective correction of pixel misalignment and color distortions.
By leveraging the pretrained visual priors of a frozen MLLM and training only the lightweight correction network, the view enhancement branch can be effectively optimized using a few paired clean views.
In this way, the view enhancement branch reduces structural and color distortions in the restored images, providing more reliable pseudo-GT supervision for subsequent 3DGS optimization.

After obtaining reliable pseudo-GT views, we further focus on reducing the computational and memory overhead of 3DGS during scene reconstruction.
To this end, we design a fast 3D scene reconstruction branch and accelerate the optimization process from three aspects: rasterization, memory layout, and parameter updates.
Specifically, we adopt opacity-aware axis-aligned rectangular bounding boxes to reduce unnecessary rasterization computations.
We reorder active Gaussians using Morton codes to improve memory-access efficiency for spatially adjacent Gaussians, and fuse the Adam updates of different parameter groups into a single CUDA kernel to reduce kernel launches and global memory access. 
With these designs, MLLM-GS significantly improves the optimization efficiency of 3DGS while maintaining high reconstruction quality.
Extensive experiments demonstrate that MLLM-GS consistently achieves high-quality NVS under diverse degradations. 
Compared with baselines trained for 30K iterations, MLLM-GS achieves superior reconstruction quality with only 5K iterations, reducing training time by approximately 9× while consuming substantially less GPU memory.

Our main contributions can be summarized as:
\begin{itemize}
    \item We propose MLLM-GS, a unified framework for fast NVS in extreme environments. 
    It combines MLLM prior-guided view enhancement with fast 3D reconstruction to enable high-quality NVS across diverse severe degradations using only a few paired clean views.

    \item We develop an MLLM prior-guided view enhancement branch that leverages the visual priors of a frozen MLLM to restore degraded views. 
    A lightweight correction network reduces residual structural and color errors under few-shot supervision, providing reliable pseudo-GT views for 3D reconstruction.

    \item We design a fast 3D scene reconstruction branch that improves 3DGS efficiency by optimizing rasterization, memory layout, and parameter updates, enabling faster scene reconstruction in degraded scenes.

    \item We conduct extensive experiments across multiple severe degradation scenarios, including smoke, low light, motion blur, and exposure variations.
    Experimental results demonstrate that MLLM-GS achieves high-quality NVS while substantially reducing reconstruction time and GPU memory consumption compared with representative methods.
\end{itemize}

\section{Related Work}

\subsection{NVS in Degraded Scenes}
3DGS~\citep{10521791,ren2026fastgs,zhang2024gaussian,sabour2025spotlesssplats,wang2025desplat,lee2026optimized,11106187,11301949,11481577} has become an effective representation for 3D reconstruction and NVS, achieving high-quality rendering with relatively low computational cost.
However, standard 3DGS~\citep{kerbl20233d} relies on reliable multi-view observations, and its performance can degrade substantially under severe imaging conditions, such as smoke, low light, and motion blur~\citep{liu2026ntire,guo2026dark3r}. 
To address this limitation, recent studies~\citep{yoon2026resplat,lin2025hqgs,liu2025deraings,lee2024deblurring,yu2026dcrr,liu2025spatiotemporal,yan2026eretinexgs} have introduced physical priors into 3DGS to improve reconstruction from degraded observations.
For example, DehazeGS~\citep{yu2026dehazegs} integrated an atmospheric scattering model into Gaussian primitives, and jointly estimated transmission and atmospheric light to recover clear scene appearances from hazy images. 
RestorGS~\citep{qiao2025restorgs} decoupled scene appearance from degradation effects and introduced depth-guided self-supervision to improve reconstruction quality.
WaterSplatting~\citep{li2025watersplatting} extended Gaussian representations with a volumetric medium field to model underwater backscattering effects.

Although these methods improve reconstruction under specific degradation conditions, they typically rely on degradation-specific imaging models and assumptions. 
Since different degradations arise from distinct imaging mechanisms, these assumptions may not hold under unknown or mixed degradations, limiting their generalization across diverse extreme environments. 
In contrast, our method leverages the general visual priors of a pretrained MLLM to enhance observations under different degradation conditions without requiring degradation-specific imaging models, thereby providing more reliable supervision for subsequent 3D reconstruction.

\subsection{Generative Priors for Restoration and Reconstruction}
Recent advances in generative modeling~\citep{conde2024instructir,lin2024diffbir,yilmaz2026edit2restore,fei2026generative,gu2026boosting,11297843,11417902,wang2026residual} have opened new opportunities for image restoration. %{wu2024waveformer,11646986,11644855,yue2025text,wu2025bvinet,zhang2023pfta,chen2025prompt} 
Unlike conventional restoration networks~\citep{11202372,gao2026airpnet,11123156,10990319,10962317,cui2026visual,li2025diffusion} that often specialize in specific degradation types, generative models leverage rich visual priors to handle diverse degradations. 
InstructIR~\citep{conde2024instructir} introduced instruction-conditioned restoration for multiple degradation tasks. 
DiffBIR~\citep{lin2024diffbir} employed diffusion priors to recover realistic image details, while Edit2Restore~\citep{yilmaz2026edit2restore} explored adapting image editing models to restoration tasks.

Recent studies~\citep{11433514,li2026hybrid,zhu2025intelligent,jiang2026multi} further demonstrated the potential of general generative models for image restoration across diverse degradations.
Such restoration strategies have also been explored for 3D reconstruction. 
For example, Smoke-GS~\citep{zheng20263d} leveraged generative visual priors to enhance degraded observations for subsequent reconstruction. 
GenSmoke-GS~\citep{cao2026gensmoke} and Dehaze-then-Splat~\citep{chen2026dehaze} investigated restoration-assisted reconstruction in smoky scenes, while NAKA-GS~\citep{zhu2026naka} addressed reconstruction under low-light conditions. 
However, generative restoration models may introduce pixel misalignment and color deviations~\citep{sun2026can,zhao2026mitigating}, which can propagate into 3D reconstruction and compromise geometry and appearance consistency~\citep{liu2026ntire,wang2026dualsplat}. 
To address this limitation, our method combines the visual priors of a frozen MLLM with a lightweight correction network trained using only a few paired clean views, providing more reliable pseudo-GT supervision for 3D reconstruction.

\subsection{Efficient 3DGS}
{Although 3DGS enables efficient rendering, its adaptive density control can introduce a large number of Gaussian primitives, resulting in substantial computational and memory overhead. 
To improve efficiency, existing studies~\citep{chen2026survey,11441989,ren2026fastgs,hahlbohm2026faster} have explored reducing redundant Gaussian growth, accelerating rasterization, and optimizing the training process.
For instance, Mini-Splatting~\citep{fang2024mini} introduced spatially aware densification and Gaussian simplification to concentrate primitives in informative regions.
Taming 3DGS~\citep{mallick2024taming} combined gradient information with auxiliary signals to guide Gaussian growth and reduce redundant densification. 
LightGaussian~\citep{fan2024lightgaussian} reduced the number of primitives through opacity- and volume-based pruning, while PUP-3DGS~\citep{hanson2025pup} employed sensitivity scores to identify and remove redundant components from pretrained representations. 
Beyond Gaussian reduction, Speedy-Splat~\citep{hanson2025speedy} improved rasterization efficiency by tightening Gaussian tile coverage to avoid unnecessary computations. 
Faster-GS~\citep{hahlbohm2026faster} further accelerated 3DGS optimization through efficient backward propagation, locality-preserving Gaussian reordering, and fused parameter updates.
Other studies~\citep{lan20253dgs2,pehlivan2026matrix,ye2025gsplat} further explored efficient optimization and GPU implementations to reduce the computational cost of 3DGS.

Despite these advances, existing acceleration methods have primarily been developed and evaluated using high-quality observations. 
In degraded scenes, residual artifacts and cross-view inconsistencies may persist even after view enhancement, potentially disrupting Gaussian optimization and inducing redundant densification.
This can further increase rasterization costs and memory-access overhead during reconstruction. 
However, the effectiveness of existing acceleration strategies under such conditions remains insufficiently explored.
To address this challenge, we investigate efficient 3DGS optimization in degraded scenes by optimizing rasterization, memory layout, and parameter updates to reduce redundant computation and memory-access overhead.

\begin{figure*}[!h]
\centering
\includegraphics[width=1\linewidth]{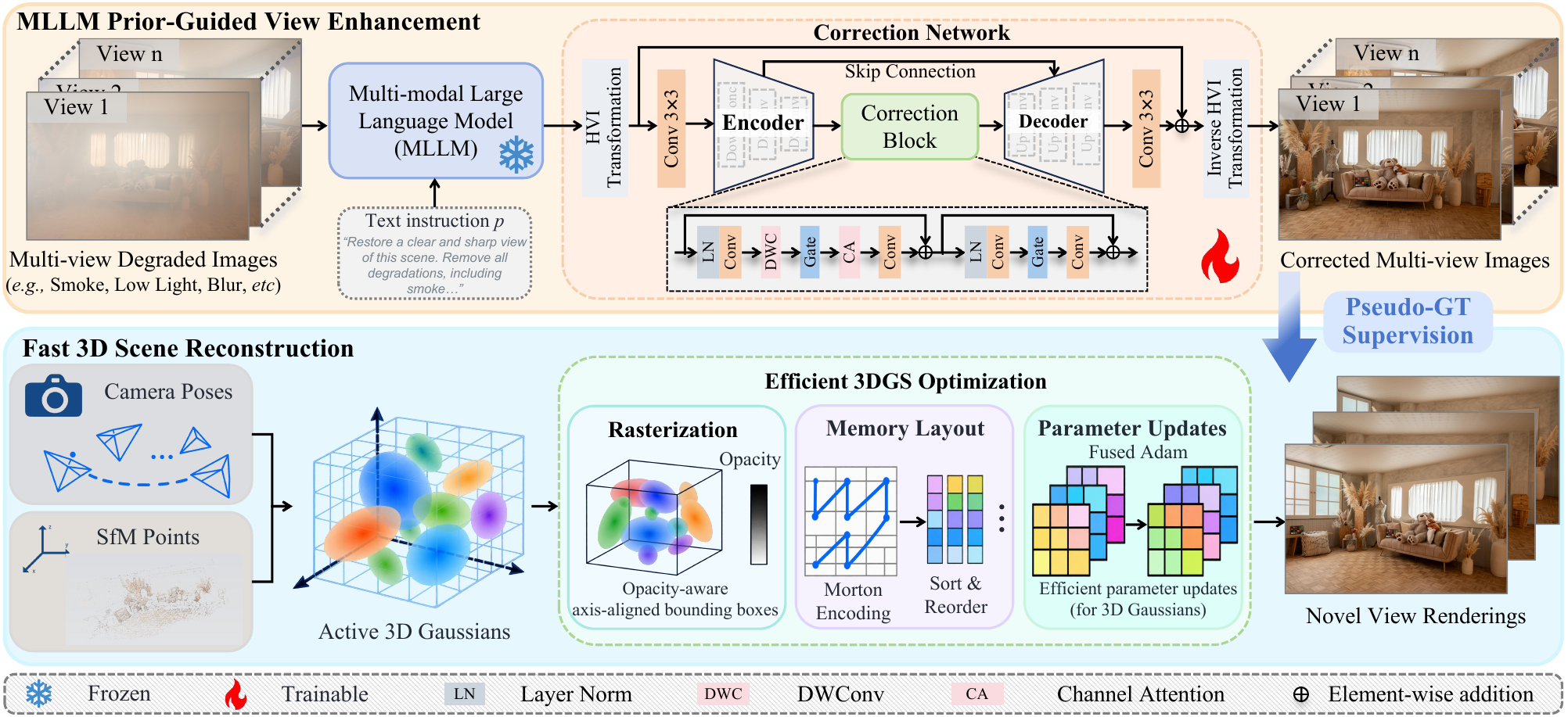}
\vspace{-0.45cm}
\caption{\textbf{Overall pipeline of the proposed MLLM-GS framework}. 
It consists of two branches: MLLM prior-guided view enhancement and fast 3D scene reconstruction.
The former combines a frozen MLLM with a lightweight correction network trained using a few paired clean views to generate reliable pseudo-GT views for subsequent 3D reconstruction.
The latter accelerates 3DGS optimization through efficient rasterization, memory layout, and parameter updates, enabling fast and high-quality NVS.}
\label{fig:method}
\end{figure*}

\section{Methodology}

\subsection{Problem Formulation}
Let $\textbf{\emph{X}}=\{\textbf{\emph{x}}_1, \textbf{\emph{x}}_2,\dots,\textbf{\emph{x}}_n\}$ denote degraded images (\emph{e.g.}, smoke, low light, defocus blur, motion blur, exposure variation) of a scene captured from $n$ different views,
where $\textbf{\emph{x}}_i\in\mathbb{R}^{h\times w\times3}$.
The corresponding camera poses are denoted by $\boldsymbol{\Pi}=\{\boldsymbol{\pi}_1,\boldsymbol{\pi}_2,\dots,\boldsymbol{\pi}_n\}$.
We assume that only a small subset of degraded views has corresponding clean images. 
Without loss of generality, we denote these paired clean images as $\textbf{\emph{C}}=\{\textbf{\emph{c}}_1, \textbf{\emph{c}}_2,\dots,\textbf{\emph{c}}_m\}$.
They correspond to $\{\textbf{\emph{x}}_1, \textbf{\emph{x}}_2,\dots,\textbf{\emph{x}}_m\}$, where $m\ll n$.
Our goal is to learn a reliable 3D scene representation $\mathcal{G}_{\theta}$ from the degraded observations $\textbf{\emph{X}}$ using only a few paired clean images for supervision, and to synthesize a clean novel view at an unseen camera pose $\bm{\pi}_t \notin \bm{\Pi}$:

\begin{equation}
\widehat{\textbf{\emph{y}}}_t = \mathcal{G}_{\theta}(\bm{\pi}_t)
\in \mathbb{R}^{h \times w \times 3}.
\label{eq:render}
\end{equation}

The rendered view $\widehat{\textbf{\emph{y}}}_t$ should be consistent with the corresponding clean ground truth image $\textbf{\emph{y}}_t\in\mathbb{R}^{h\times w\times3}$.
It should faithfully reflect the scene geometry and view-dependent appearance, and preserve realistic textures and colors.

To achieve this goal, we propose MLLM-GS, a framework for fast NVS in extreme environments.
As illustrated in Fig.~\ref{fig:method}, MLLM-GS consists of two branches: MLLM prior-guided view enhancement (Section~\ref{MLLM Prior-Guided Image Enhancement}) and fast 3D scene reconstruction (Section~\ref{Fast 3D Scene Reconstruction}).
The former leverages the rich visual priors of a frozen MLLM to enhance degraded observations and employs a lightweight correction network to reduce pixel misalignment and color distortions under few-shot supervision. 
This branch generates reliable pseudo-GT views for all degraded observations, providing supervision for subsequent 3D reconstruction.
The latter reconstructs the 3D scene using these pseudo-GT views and improves 3DGS optimization efficiency in terms of rasterization, memory layout, and parameter updates.
By combining reliable view enhancement with efficient 3D reconstruction, MLLM-GS enables high-quality and fast NVS across diverse severe degradations.

\subsection{MLLM Prior-Guided View Enhancement}
\label{MLLM Prior-Guided Image Enhancement}
Images captured in extreme environments often suffer from severe degradations, including smoke, low light, defocus blur, motion blur, and exposure variation. 
These degradations can damage both structural and color information.
Directly using degraded views to supervise 3D reconstruction can introduce errors in both scene geometry and appearance, leading to suboptimal NVS results.
An intuitive solution is to restore the views before reconstruction.
However, existing restoration networks~\citep{11202372,11123156} typically require extensive paired training data and are often specialized for specific degradation types, limiting their applicability under diverse degradations.
To address these limitations, we develop an MLLM prior-guided view enhancement branch that combines a frozen MLLM with a lightweight correction network. 
The MLLM leverages general visual priors to enhance observations under diverse degradations, while the correction network reduces residual structural and color distortions using only a few paired clean views. 
In this way, the branch extends limited clean supervision to all input views, providing reliable pseudo-GT images for subsequent 3D reconstruction.

Specifically, we first feed each degraded view $\textbf{\emph{x}}_i$ and a text instruction $\textbf{\emph{p}}$ into a frozen MLLM $\mathcal{M}(\cdot)$ to obtain an enhanced image $\textbf{\emph{e}}_i$:
\begin{equation}
\textbf{\emph{e}}_i = \mathcal{M}(\textbf{\emph{x}}_i,\textbf{\emph{p}}).
\end{equation}

Although $\textbf{\emph{e}}_i$ recovers plausible scene structures and textures, it may still contain residual structural and color distortions.
To reduce these errors, we further process $\textbf{\emph{e}}_i$ using the correction network illustrated in Fig.~\ref{fig:method}. 
The correction network first converts $\textbf{\emph{e}}_i$ into the Horizontal/Vertical-Intensity (HVI) color space~\citep{yan2025hvi}. 
In RGB space, luminance and chrominance information are coupled, making it difficult to correct structural and color distortions separately.
In contrast, the HVI color space separates luminance from chromatic information, allowing the network to model structural and color deviations more effectively. 
The transformation is formulated as
\begin{equation}
\textbf{\emph{s}}_i = \mathcal{T}(\textbf{\emph{e}}_i),
\end{equation}
where $\mathcal{T}(\cdot)$ denotes the HVI transform and
$\textbf{\emph{s}}_i= [\textbf{\emph{h}}_i,\textbf{\emph{v}}_i,\textbf{\emph{d}}_i] \in \mathbb{R}^{h\times w\times 3}$ is the HVI representation of $\textbf{\emph{e}}_i$.
$\textbf{\emph{h}}_i$ and $\textbf{\emph{v}}_i$ encode the chromatic information and $\textbf{\emph{d}}_i$ represents the luminance information.

Subsequently, the HVI representation $\textbf{\emph{s}}_i$ is projected into a high-dimensional feature space through a convolutional layer and processed by a U-Net to correct residual structural and color distortions.
The encoder progressively extracts multi-scale features to capture both local details and contextual information. 
At the bottleneck, a correction block integrates depthwise convolution, a gating mechanism, and channel attention to further refine the encoded features.
Depthwise convolution captures local spatial information, while the gating mechanism adaptively emphasizes features relevant to distortion correction.
Channel attention further recalibrates feature responses across channels to improve the modeling of structural and color information.
The decoder progressively restores the spatial resolution and integrates encoder features through skip connections to preserve fine-grained details.
Finally, the corrected HVI representation $\widehat{\textbf{\emph{s}}}_i$ is converted back to RGB space through the inverse HVI transformation, yielding the corrected view $\widehat{\textbf{\emph{c}}_i}$:
\begin{equation}
\widehat{\textbf{\emph{c}}_i} = \mathcal{T}^{-1}(\widehat{\textbf{\emph{s}}}_i).
\end{equation}

During training, the MLLM remains frozen, and only the lightweight correction network with approximately 4M parameters is optimized. 
By combining the pretrained visual priors of the frozen MLLM with few-shot supervised correction, the proposed branch can effectively refine the enhanced views using only a few paired clean images.
The training objective combines a Charbonnier loss $\mathcal{L}_{\text{Char}}(\cdot,\cdot)$ and a frequency-domain loss $\mathcal{L}_{\text{Freq}}(\cdot,\cdot)$. 
The former minimizes pixel-level reconstruction errors, while the latter encourages the preservation of image details in the frequency domain. 
The overall loss is defined as
\begin{equation}
\mathcal{L}_{\mathrm{en}} = \mathcal{L}_{\text{Char}}(\widehat{\textbf{\emph{c}}_i},\textbf{\emph{c}}_i)+ \lambda \mathcal{L}_{\text{Freq}}(\widehat{\textbf{\emph{c}}_i},\textbf{\emph{c}}_i),
\end{equation}
where $\textbf{\emph{c}}_i$ ($i\in\{1,\cdots,m\}$) denotes the corresponding clean ground-truth image and $\lambda$ balances the two loss terms.
After training, the enhancement branch is applied to all degraded observations to generate corrected views, which serve as pseudo-GT supervision for subsequent 3D reconstruction.

\subsection{Fast 3D Scene Reconstruction}
\label{Fast 3D Scene Reconstruction}

After view enhancement, the corrected views serve as pseudo-GT supervision for 3D scene reconstruction. 
However, residual artifacts and cross-view inconsistencies may still affect Gaussian optimization and induce redundant densification, increasing the computational and memory overhead of 3DGS.
First, the conservative screen-space bounding boxes used in standard 3DGS include regions with negligible Gaussian contributions, resulting in unnecessary rasterization operations.
Second, adaptive densification and pruning can scatter spatially neighboring Gaussians across memory, reducing memory-access efficiency. 
Third, updating different Gaussian attributes through separate optimization kernels introduces additional kernel launches and global memory accesses.
Inspired by Faster-GS~\citep{hahlbohm2026faster}, we develop a fast 3D scene reconstruction branch that accelerates 3DGS optimization from three aspects: rasterization, memory layout, and parameter updates.

\subsubsection{Rasterization}
Standard 3DGS assigns a conservative square bounding box to each projected Gaussian, with its half-extent determined by the larger eigenvalue of the projected 2D covariance matrix. 
However, this bounding box can substantially overestimate the effective screen-space coverage, especially for anisotropic Gaussians. 
In degraded scenes, residual artifacts and cross-view inconsistencies in the enhanced views may induce redundant Gaussian densification, further increasing rasterization overhead. 
To reduce unnecessary computation, we adopt opacity-aware, axis-aligned rectangular bounding boxes that estimate the effective Gaussian extent along each screen axis separately.

Specifically, let $o$ denote the opacity of a Gaussian primitive and ${\Sigma}_{\text{2D}}$ denote its projected 2D covariance matrix.
Given an opacity truncation threshold $\tau_\alpha$, the half-extents along the horizontal and vertical directions are defined as
\begin{equation}
\begin{cases}
r_x = \sqrt{2\ln(o / \tau_\alpha) \cdot \Sigma_{\text{2D}}^{(1,1)}},\\
r_y = \sqrt{2\ln(o / \tau_\alpha) \cdot \Sigma_{\text{2D}}^{(2,2)}},
\end{cases}
\label{eq:bbox_half_extent}
\end{equation}
where $\tau_\alpha = 1/255$.
$\Sigma_{\text{2D}}^{(1,1)}$ and $\Sigma_{\text{2D}}^{(2,2)}$ represent the projected variances along the horizontal and vertical screen axes, respectively.
Gaussians with $o\leq\tau_\alpha$ can be excluded because their maximum opacity contribution does not exceed the truncation threshold. 
By incorporating opacity into the extent estimation, the bounding region adapts to both the spatial distribution and opacity of each Gaussian. 
By excluding regions with negligible Gaussian contributions, this strategy reduces unnecessary Gaussian–tile intersections, sorting operations, and forward rasterization costs, thereby improving reconstruction efficiency.

Although the opacity-aware bounding boxes reduce redundant computations in the forward pass, gradient computation during backward propagation remains computationally expensive. 
In standard 3DGS, pixel threads accumulate Gaussian gradients through global-memory atomic operations, which can cause contention and repeated memory accesses when numerous Gaussians contribute to the same tiles.
To reduce this overhead, we adopt a bucket-based backward propagation strategy. For a tile containing $K$ sorted Gaussians, we partition them into buckets of size $B=32$. 
The accumulated color and transmittance at the boundary of bucket \(b\) are defined as
\begin{equation}
\begin{cases}
\mathbf{Q}_b(p) = \sum_{k=1}^{bB} \alpha_k(p) T_{k-1}(p)\emph{\textbf{f}}_k,\\
U_b(p) = \prod_{k=1}^{bB}(1-\alpha_k(p)),
\end{cases}
\end{equation}
where $p$ is a pixel, $b$ indexes the buckets, $\alpha_{k}(p)$ is the opacity contribution of the $k$-th Gaussian at pixel $p$, and $\emph{\textbf{f}}_k$ is its color.
The accumulated transmittance before the $k$-th Gaussian is $T_{k-1}(p)=\prod_{j=1}^{k-1}(1-\alpha_j(p))$, with $T_{0}(p)=1$.
These accumulated states provide the color and transmittance information at bucket boundaries.
During backward propagation, each warp uses the corresponding boundary states to reconstruct the blending process within a bucket, avoiding repeated traversal of all preceding Gaussians. 
This strategy reduces redundant gradient computations and global-memory accesses, improving backward propagation efficiency while preserving the original blending formulation.

\subsubsection{Memory Layout}
The adaptive density-control mechanism in 3DGS periodically clones, splits, and prunes Gaussian primitives. 
Newly generated Gaussians are typically appended to parameter buffers, causing spatially neighboring Gaussians to become scattered across memory as optimization proceeds.
Consequently, processing nearby Gaussians may require accessing distant memory locations, reducing cache efficiency and increasing memory-access overhead. 
To improve memory locality, we periodically reorder Gaussian primitives using Morton encoding, which maps three-dimensional positions into a one-dimensional Z-order sequence.

For each Gaussian mean $\boldsymbol\mu = (x, y, z)$, we compute a Morton code as
\begin{equation}
W(\boldsymbol\mu) = \operatorname{OR}_{k=0}^{2} \left( \phi\left( \left\lfloor 1023 \cdot \frac{\mu_k - \text{min}_k}{\text{max}_k - \text{min}_k} \right\rfloor \right) \ll k \right),
\label{eq:morton_code}
\end{equation}
where $\operatorname{OR}_{k=0}^{2}$ denotes the bitwise OR operation, and $\phi(\cdot)$ represents bit expansion for Morton encoding, and $\ll$ denotes the bitwise left-shift operation.
Each coordinate is normalized according to the scene bounds and quantized into a 10-bit integer. 
The bits of the three coordinates are then interleaved to generate a Morton code.

Morton encoding approximately preserves spatial locality by assigning nearby 3D positions to nearby locations in the one-dimensional ordering. 
We apply radix sorting to the Morton codes and reorder the corresponding Gaussian parameter tensors according to the sorted indices.
This organization improves the spatial locality of Gaussian memory accesses, allowing neighboring threads to access more contiguous memory regions during optimization. 
To limit the additional sorting cost, reordering is performed only at densification intervals rather than at every training iteration.

\subsubsection{Parameter Updates}
Standard 3DGS~\citep{kerbl20233d} optimizes Gaussian attributes, including position, rotation, scale, opacity, and spherical harmonic coefficients, using Adam~\citep{kingma2014adam}.
These attributes are organized into separate parameter groups, and their updates typically involve multiple kernel launches and global-memory accesses. 
Each update requires reading gradients, optimizer states, and parameter values from memory and writing the updated results back. 
As the number of Gaussian primitives increases, these operations introduce considerable computational and memory-bandwidth overhead. 
Inspired by Faster-GS~\citep{hahlbohm2026faster}, we fuse the Adam updates of different parameter groups into a single CUDA kernel and use contiguous buffers for unified parameter updates, reducing the overhead.

Specifically, we fuse the Adam updates of different Gaussian parameter groups into a single CUDA kernel and organize the corresponding parameters and optimizer states in contiguous memory buffers. 
Each thread loads the parameter value, gradient, and moment estimates into registers, where the Adam moment updates, bias correction, and parameter update are performed using fused arithmetic operations. 
The updated parameters and optimizer states are then written back to global memory. 
By reducing kernel launches and redundant global-memory accesses, this strategy improves the efficiency of Gaussian parameter updates, particularly when optimizing a large number of Gaussian primitives.

For the fast 3D scene reconstruction branch, we use the enhanced views generated by the MLLM prior-guided view enhancement branch as pseudo-GT to supervise 3DGS.
The training objective combines an $\ell_1$ loss $\mathcal{L}_{1}(\cdot,\cdot)$ and a D-SSIM loss $\mathcal{L}_{\text{D-SSIM}}(\cdot,\cdot)$ to constrain both pixel values and structural information:
\begin{equation}
\mathcal{L}_{3D}=(1-\beta)\,\mathcal{L}_{1}(\widehat{\bm{y}}_i,\widehat{\bm{c}}_i)+\beta\,\mathcal{L}_{\text{D-SSIM}}(\widehat{\bm{y}}_i,\widehat{\bm{c}}_i),
\label{eq:3d}
\end{equation}
where $\widehat{\mathbf{y}}_i$ denotes the rendered image at the $i$-th training viewpoint.
$\widehat{\bm{c}}_i$ represents the corresponding corrected pseudo-GT view, and $\beta$ is a balancing parameter.

\section{Experiments}

\subsection{Experimental Setting}
\textbf{Implementation Details:}
{
For the MLLM prior-guided view enhancement branch, we use Nano Banana Pro~\citep{raisinghani2025nano} as the default frozen MLLM.
For the correction network, the MLLM-enhanced images are first converted into HVI color space using a learnable density parameter initialized to $0.2$.
The correction network then refines the HVI representations, which are converted back to RGB space to obtain the corrected images.
We adopt the AdamW~\citep{loshchilov2017decoupled} optimizer with an initial learning rate of $1\times10^{-4}$, coupled with a cosine annealing schedule~\citep{loshchilov2016sgdr} for dynamic learning rate adjustment, and train on randomly cropped $256\times256$ patches.
The correction network is trained using only the first three pairs of degraded and corresponding clean images from each scene. 
}
\begin{figure}[t!]
\centering
\includegraphics[width=0.9\linewidth]{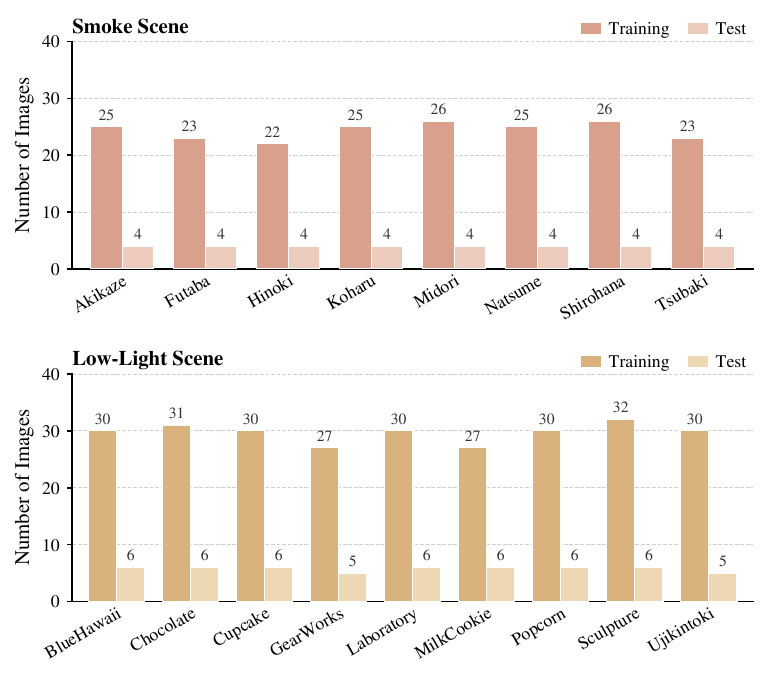}
\vspace{-0.7em}
\caption{The view statistics of smoke and low-light scenes from the RealX3D benchmark.
}
\label{fig:dataset_stat}
%\vspace{-1.0em}
\vspace{-1em}
\end{figure}

For the fast 3D scene reconstruction branch, we build upon the 3D Gaussian Splatting framework~\citep{kerbl20233d} and implement it with the gsplat library~\citep{ye2025gsplat}. 
Each scene is optimized for $30,000$ iterations.
The training objective combines a reconstruction loss $\gL_1$ and a D-SSIM loss $\gL_{\text{D-SSIM}}$. The balance weight $\beta$ is set to 0.2. 
% $0.8$ and $0.2$, respectively.
Optimization is carried out using the fused Adam optimizer with $\epsilon = 10^{-15}$. 
The learning rates for the Gaussian scales, rotations, and opacities are set to $5\times10^{-3}$, $1\times10^{-3}$, and $5\times10^{-2}$, respectively. 
For the spherical harmonics coefficients, the learning rate is $2.5\times10^{-3}$ for the DC component and $1.25\times10^{-4}$ for the higher-order components.

\begin{table*}[t]
\centering
\setlength{\tabcolsep}{5pt}
\caption{Quantitative evaluation of 3D reconstruction methods on the smoke scene of the RealX3D benchmark.
Best results are highlighted in dark gray bold, and second-best in light gray.}
\label{tab:smoke_exp}
\resizebox{1.0\linewidth}{!}{
\begin{tabular}{|c|c|cccccccc|c|}
\hline\thickhline
Method & Metrics & Akikaze & Futaba & Hinoki & Koharu & Midori & Natsume & Shirohana & Tsubaki & Avg. \\
\hline

\multirow{3}{*}{3DGS~\citep{kerbl20233d}}
& PSNR $\uparrow$ & 10.80 & 13.48 & 7.47 & 11.87 & 13.84 & 9.74 & 8.91 & 15.41 & 11.44 \\
& SSIM $\uparrow$ & 0.548 & 0.695 & 0.315 & 0.611 & 0.700 & 0.604 & 0.419 & 0.739 & 0.579 \\
& LPIPS $\downarrow$ & 0.659 & 0.619 & 0.683 & 0.573 & 0.548 & 0.633 & 0.749 & 0.583 & 0.631 \\
\hline

\multirow{3}{*}{SeaSplat~\citep{yang2025seasplat}}
& PSNR $\uparrow$ & 11.21 & 8.11 & 13.23 & 9.98 & 5.04 & 8.49 & 9.21 & 8.95 & 9.28 \\
& SSIM $\uparrow$ & 0.497 & 0.497 & 0.422 & 0.500 & 0.330 & 0.463 & 0.347 & 0.507 & 0.445 \\
& LPIPS $\downarrow$ & 0.863 & 0.727 & 0.677 & 0.619 & 0.850 & 0.863 & 0.847 & 0.637 & 0.760 \\
\hline

\multirow{3}{*}{WaterSplatting~\citep{li2025watersplatting}}
& PSNR $\uparrow$ & 9.54 & 8.64 & 14.06 & 14.25 & 4.64 & 7.82 & 8.22 & 7.24 & 9.30 \\
& SSIM $\uparrow$ & 0.440 & 0.522 & 0.432 & 0.644 & 0.286 & 0.373 & 0.338 & 0.375 & 0.426 \\
& LPIPS $\downarrow$ & 0.753 & 0.674 & 0.827 & 0.641 & 0.778 & 0.691 & 0.704 & 0.638 & 0.713 \\
\hline

\multirow{3}{*}{I2-NeRF~\citep{liu2026i2}}
& PSNR $\uparrow$ & 7.19 & 5.46 & 13.35 & 9.62 & 3.52 & 6.13 & 5.70 & 6.14 & 7.14 \\
& SSIM $\uparrow$ & 0.252 & 0.252 & 0.405 & 0.402 & 0.150 & 0.216 & 0.138 & 0.227 & 0.255 \\
& LPIPS $\downarrow$ & 0.740 & 0.696 & 0.657 & 0.640 & 0.745 & 0.706 & 0.750 & 0.697 & 0.704 \\
\hline

\multirow{3}{*}{SeaThruNeRF~\citep{levy2023seathru}}
& PSNR $\uparrow$ & 8.17 & 10.55 & 4.90 & 7.77 & 12.37 & 8.48 & 8.43 & 11.43 & 9.01 \\
& SSIM $\uparrow$ & 0.513 & 0.666 & 0.263 & 0.539 & 0.687 & 0.595 & 0.412 & 0.702 & 0.547 \\
& LPIPS $\downarrow$ & 0.672 & 0.620 & 0.673 & 0.653 & 0.575 & 0.613 & 0.739 & 0.597 & 0.643 \\
\hline

\multirow{3}{*}{\shortstack{ConvIR-B~\citep{cui2024revitalizing} \\ +  3DGS~\citep{kerbl20233d}}}
& PSNR $\uparrow$ & 15.00 & 16.61 & 15.15 & 16.97 & 15.83 & 13.52 & 15.55 & 17.31 & 15.74 \\
& SSIM $\uparrow$ & 0.616 & 0.732 & 0.489 & 0.652 & 0.716 & 0.669 & 0.522 & 0.776 & 0.647 \\
& LPIPS $\downarrow$ & 0.846 & 0.682 & 0.895 & 0.625 & 0.614 & 0.795 & 0.767 & 0.671 & 0.737 \\
\hline

\multirow{3}{*}{\shortstack{MB-TF-B V2~\citep{10962317} \\  + 3DGS~\citep{kerbl20233d}}}
& PSNR $\uparrow$  & 18.68 & 13.42 & 16.75 & 18.28 & 17.90 & 14.87 & 16.39 & 18.01 & 16.79 \\
& SSIM $\uparrow$  & 0.666 & 0.685 & 0.513 & 0.722 & 0.729 & 0.663 & 0.562 & 0.764 & 0.663 \\
& LPIPS $\downarrow$ & 0.560 & 0.646 & 0.762 & 0.483 & 0.610 & 0.680 & 0.636 & 0.611 & 0.624  \\
\hline

\multirow{3}{*}{\shortstack{ConvIR-B~\citep{cui2024revitalizing} \\  + WaterSplatting~\citep{li2025watersplatting}}}
& PSNR $\uparrow$ & 15.49 & 17.18 & 15.67 & 17.31 & 16.67 & 15.20 & 16.18 & 18.36 & 16.51  \\
& SSIM $\uparrow$  & 0.657 & 0.742 & 0.507 & 0.689 & 0.710 & 0.709 & 0.571 & 0.759 & 0.668 \\
& LPIPS $\downarrow$ & 0.651 & 0.567 & 0.661 & 0.492 & 0.500 & 0.633 & 0.733 & 0.524 & 0.595 \\
\hline

\multirow{3}{*}{\shortstack{MB-TF-B V2~\citep{10962317}\\  + WaterSplatting~\citep{li2025watersplatting}}}
& PSNR $\uparrow$  & 19.55 & 13.52 & 16.99 & 18.59 & 19.60 & 16.91 & 16.73 & 18.74 & 17.58 \\
& SSIM $\uparrow$  & \cellcolor{gray!25}0.698 & 0.672 & 0.496 & 0.704 & 0.705 & 0.722 & 0.541 & 0.780 & 0.665  \\
& LPIPS $\downarrow$ & \cellcolor{gray!25}0.323 & 0.579 & 0.684 & 0.397 & 0.459 & 0.547 & \cellcolor{gray!25}0.534 & 0.533 & 0.507 \\
\hline

\multirow{3}{*}{Dehaze-then-Splat~\citep{chen2026dehaze}}
& PSNR $\uparrow$ & \cellcolor{gray!25}20.98 & 18.05 & 18.20 & 18.67 & 18.12 & 15.09 & 15.43 & \cellcolor{gray!25}21.66 & 18.28 \\
& SSIM $\uparrow$ & 0.683 & 0.705 & 0.507 & 0.624 & 0.711 & 0.634 & 0.448 & 0.699 & 0.626 \\
& LPIPS $\downarrow$ & 0.600 & 0.596 & 0.676 & 0.510 & 0.578 & 0.623 & 0.652 & 0.583 & 0.602 \\
\hline

 \multirow{3}{*}{Smoke-GS~\citep{zheng20263d}}
 & PSNR $\uparrow$ & - & 16.30 & 17.01 & 17.57 &19.12 &20.59 &17.59 &22.48  & 18.67\\
 & SSIM $\uparrow$ & - & 0.721 & 0.485 & 0.685 &0.723 &0.716 &\cellcolor{gray!25}0.579 &0.785  &\cellcolor{gray!25} 0.670 \\
 & LPIPS $\downarrow$ & - & 0.460 & 0.579 & 0.438 &\cellcolor{gray!25}0.321 &\cellcolor{gray!25}0.446 &0.544 &0.361  & 0.449\\
 \hline

\multirow{3}{*}{GenSmoke-GS~\citep{cao2026gensmoke}}
& PSNR $\uparrow$ & - & \cellcolor{gray!25}20.97 & \cellcolor{gray!25}19.12 & \cellcolor{gray!25}20.87 & \cellcolor{gray!25}20.56 & \cellcolor{gray!25}20.76 & \cellcolor{gray!25}17.50 & \cellcolor{gray!25}21.66 & \cellcolor{gray!25}20.21 \\
& SSIM $\uparrow$ & - & \cellcolor{gray!40}\textbf{0.807} & \cellcolor{gray!40}\textbf{0.607} & \cellcolor{gray!25}0.783 & \cellcolor{gray!25}0.791 & \cellcolor{gray!25}0.745 & 0.569 & \cellcolor{gray!25}0.799 & \cellcolor{gray!40}\textbf{0.729} \\
& LPIPS $\downarrow$ & - & \cellcolor{gray!25}0.381 & \cellcolor{gray!25}0.526 & \cellcolor{gray!25}0.388 & 0.391 & 0.451 & 0.590 & \cellcolor{gray!25}0.394 & \cellcolor{gray!25}0.446 \\
\hline

\multirow{3}{*}{\textbf{MLLM-GS (Ours)}} & PSNR $\uparrow$ & \cellcolor{gray!40}\textbf{27.44} & \cellcolor{gray!40}\textbf{22.71} & \cellcolor{gray!40}\textbf{20.15} & \cellcolor{gray!40}\textbf{22.25} & \cellcolor{gray!40}\textbf{21.21} & \cellcolor{gray!40}\textbf{21.85} & \cellcolor{gray!40}\textbf{19.16} & \cellcolor{gray!40}\textbf{23.35} 
& \cellcolor{gray!40}\textbf{22.27} \\
& SSIM $\uparrow$ 
& \cellcolor{gray!40}\textbf{0.761}
& \cellcolor{gray!25}0.767 
& \cellcolor{gray!25}0.564
& \cellcolor{gray!40}\textbf{0.793}
& \cellcolor{gray!40}\textbf{0.795}
& \cellcolor{gray!40}\textbf{0.761} 
& \cellcolor{gray!40}\textbf{0.584}
&\cellcolor{gray!40}\textbf{0.803}
& \cellcolor{gray!40}\textbf{0.729} \\
& LPIPS $\downarrow$ & \cellcolor{gray!40}\textbf{0.234} & \cellcolor{gray!40}\textbf{0.330} & \cellcolor{gray!40}\textbf{0.451} & \cellcolor{gray!40}\textbf{0.377} & \cellcolor{gray!40}\textbf{0.270} & \cellcolor{gray!40}\textbf{0.440} & \cellcolor{gray!40}\textbf{0.438} & \cellcolor{gray!40}\textbf{0.358}
& \cellcolor{gray!40}\textbf{0.362} \\
\hline
\end{tabular}}
%\vspace{-0.2cm}
\end{table*}

\textbf{Benchmarks:}
We evaluate MLLM-GS on the RealX3D benchmark~\citep{liu2026realx3d}, which provides real-world multi-view observations captured under challenging imaging conditions, including smoke, low light, and exposure variations.
RealX3D provides degraded observations and their corresponding clean reference images, enabling quantitative evaluation of NVS quality under challenging imaging conditions. 
For each scene, the training and test views are disjoint.
During training, only three paired degraded and clean views are used to supervise the correction network. 
The resulting enhancement branch generates pseudo-GT images for all training views, while the clean reference images at test viewpoints are used exclusively for evaluation. 
Fig.~\ref{fig:dataset_stat} presents the numbers of training and test views for the smoke and low-light scenes.

\textbf{Evaluation Metrics:}
To evaluate rendering quality, we report Peak Signal-to-Noise Ratio (PSNR), Structural Similarity Index Measure (SSIM)~\citep{wang2004image}, and Learned Perceptual Image Patch Similarity (LPIPS)~\citep{zhang2018unreasonable} for quantitative evaluation.
Specifically, PSNR measures pixel-level reconstruction fidelity, while SSIM evaluates structural similarity between the rendered and reference images.
Higher values  ($\uparrow$) of these two metrics indicate better reconstruction quality of the novel view.
LPIPS measures the perceptual similarity between the rendered and reference images in a deep feature space.
A lower value ($\downarrow$) indicates better perceptual quality.
In implementation, LPIPS is computed using the AlexNet~\citep{krizhevsky2012imagenet} backbone following the official code~\citep{zhang2018unreasonable}.
Furthermore, to evaluate computational efficiency and representation complexity, we additionally report GPU memory consumption (VRAM), training time (Training), and the final number of Gaussian primitives (\#Gs).

\subsection{Results in Smoke Scenes}

\textbf{Quantitative Results:}
For smoke scenes, we evaluate all methods on eight scenes from the RealX3D~\citep{liu2026realx3d} benchmark.
We compare our method with three categories of baselines:
(1) classic 3D reconstruction methods, including 3DGS~\citep{kerbl20233d}, SeaSplat~\citep{yang2025seasplat}, WaterSplatting~\citep{li2025watersplatting}, I2-NeRF~\citep{liu2026i2}, and SeaThruNeRF~\citep{levy2023seathru};
(2) restoration-based reconstruction pipelines that combine ConvIR-B~\citep{cui2024revitalizing} or MB-TF-B V2~\citep{10962317} with 3DGS or WaterSplatting;
and (3) the top-three solutions from the NTIRE 2026 3D Restoration and Reconstruction Challenge, including GenSmoke-GS~\citep{cao2026gensmoke}, Smoke-GS~\citep{zheng20263d}, and Dehaze-then-Splat~\citep{chen2026dehaze}.
For a fair comparison, all restoration-based reconstruction pipelines are trained under the same supervision setting as MLLM-GS.
ConvIR-B~\citep{cui2024revitalizing} and MB-TF-B V2~\citep{10962317} are fine-tuned using the same three paired degraded and clean views per scene, and their restored images are subsequently used to train 3DGS or WaterSplatting.
As shown in Table~\ref{tab:smoke_exp}, classic 3D reconstruction methods generally achieve lower reconstruction quality under severe smoke degradation, indicating the difficulty of recovering clear scenes directly from degraded observations. 
Restoration-based reconstruction pipelines improve reconstruction quality by enhancing the input images before 3D reconstruction, achieving average PSNR values ranging from $15.74$ dB to $17.58$ dB.
However, residual artifacts and color distortions in the enhanced views can propagate into the reconstruction process, limiting the final rendering quality.
Challenge-based methods achieve stronger performance, with average PSNR values ranging from $18.28$ dB to $20.21$ dB. 
In contrast, our method achieves the highest average PSNR of $22.27$ dB and the lowest average LPIPS of $0.362$.
Notably, under identical supervision, MLLM-GS achieves an average PSNR improvement of $4.69$ dB over the strongest restoration-based pipeline.
This indicates that the performance gains arise from the effective integration of pretrained MLLM priors and few-shot correction rather than additional supervision. 
These results further demonstrate the effectiveness of MLLM-GS in providing reliable supervision for high-quality NVS.

\begin{figure*}[!t]
\centering
\includegraphics[width=1\linewidth]{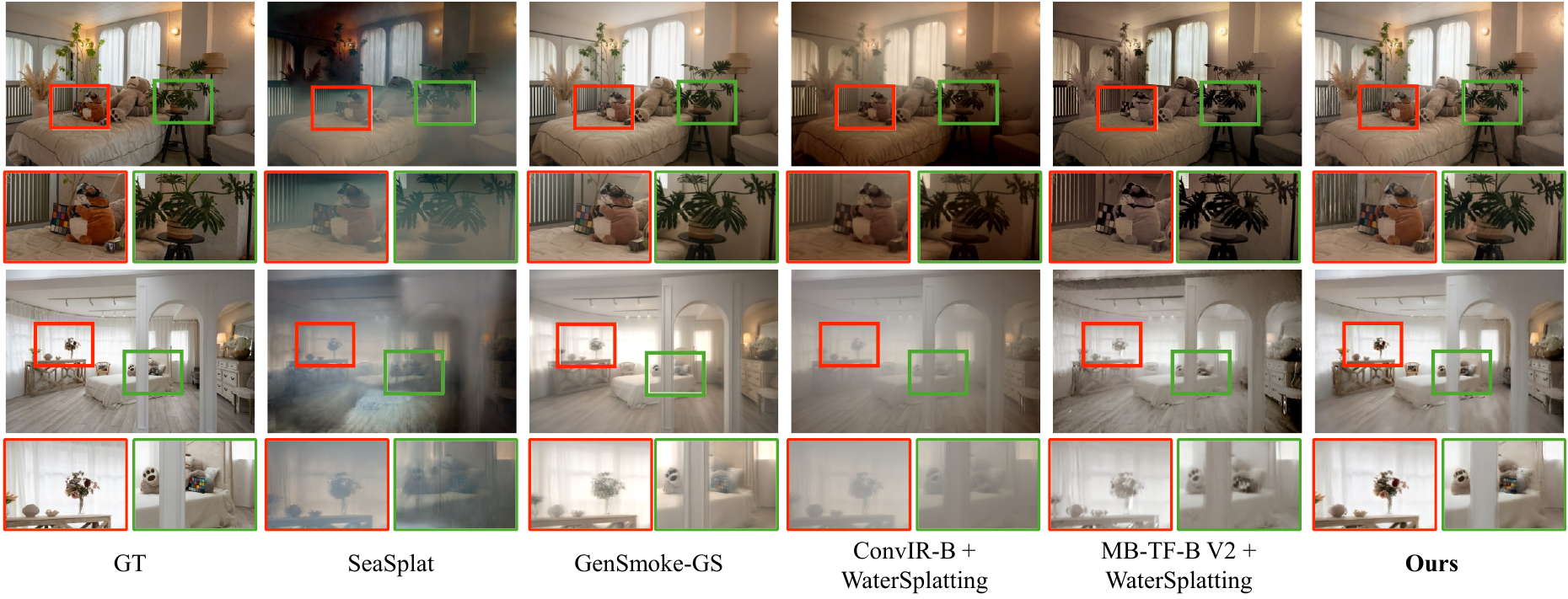}
\vspace{-0.4cm}
\caption{Qualitative comparison of NVS results under smoke scenes.}
\label{fig:vis_hazy_compare}
\vspace{-0.2cm}
\end{figure*}

\textbf{Qualitative Results:}
Fig.~\ref{fig:vis_hazy_compare} presents qualitative comparisons of NVS results on two smoke scenes.
SeaSplat~\citep{yang2025seasplat} produces renderings with substantial residual haze, low contrast, and blurred scene structures. 
Restoration-based reconstruction pipelines combining ConvIR-B~\citep{cui2024revitalizing} or MB-TF-B V2~\citep{10962317} with WaterSplatting~\citep{li2025watersplatting} improve scene visibility but still exhibit color distortions and loss of fine-grained details.
GenSmoke-GS~\citep{cao2026gensmoke} recovers clearer structures, although residual haze and color deviations remain visible in some regions. 
In contrast, our method substantially reduces residual haze while preserving scene structures, fine textures, and natural colors. 
The rendered results are visually closer to the ground-truth images, particularly in object boundaries and textured regions. 
These results demonstrate the effectiveness of MLLM-GS in generating visually faithful novel views under severe smoke degradation.

\begin{table*}[t]
\centering
\setlength{\tabcolsep}{1.5pt}
\caption{Quantitative evaluation of 3D reconstruction methods on the low-light scene of the RealX3D benchmark.
Best results are highlighted in dark gray bold, and second-best in light gray.}
\label{tab:lowlight_exp}
\resizebox{1.0\linewidth}{!}{
\begin{tabular}{|c|c|ccccccccc|c|}
\hline\thickhline
Method & Metrics & BlueHawaii & Chocolate & Cupcake & GearWorks & Laboratory & MilkCookie & Popcorn & Sculpture & Ujikintoki & Avg. \\ \hline

\multirow{3}{*}{3DGS~\citep{kerbl20233d}}
& PSNR $\uparrow$ & 7.49 & 7.89 & 4.42 & 6.90 & 7.59 & 6.15 & 7.46 & 5.85 & 6.23 & 6.66 \\
& SSIM $\uparrow$ & 0.055 & 0.084 & 0.064 & 0.081 & 0.041 & 0.053 & 0.062 & 0.019 & 0.062 & 0.058 \\
& LPIPS $\downarrow$ & 0.671 & 0.646 & 0.625 & 0.658 & 0.610 & 0.653 & 0.650 & 0.733 & 0.683 & 0.659 \\
\hline

\multirow{3}{*}{LITA-GS~\citep{zhou2025lita}}
& PSNR $\uparrow$ & 17.30 & 17.94 & 13.07 & 10.90 & 17.56 & 12.74 & 18.97 & 12.84 & 18.82 & 15.57 \\
& SSIM $\uparrow$ & 0.624 & 0.541 & 0.643 & 0.344 & 0.597 & 0.573 & 0.557 & 0.211 & 0.656 & 0.527 \\
& LPIPS $\downarrow$ & 0.546 & 0.557 & 0.326 & 0.521 & 0.435 & 0.478 & 0.448 & 0.715 & 0.497 & 0.503 \\
\hline

\multirow{3}{*}{I2-NeRF~\citep{liu2026i2}}
& PSNR $\uparrow$ & 18.08 & 19.77 & 12.68 & 12.22 & 16.34 & 14.78 & 17.08 & 9.64 & 19.02 & 15.51 \\
& SSIM $\uparrow$ & 0.657 & 0.571 & 0.608 & 0.511 & 0.630 & 0.649 & 0.543 & 0.277 & 0.666 & 0.568 \\
& LPIPS $\downarrow$ & 0.546 & 0.561 & 0.493 & 0.543 & 0.486 & 0.544 & 0.488 & 0.649 & 0.474 & 0.532 \\
\hline

\multirow{3}{*}{Aleth-NeRF~\citep{cui2024aleth}}
& PSNR $\uparrow$ & 15.84 & 11.53 & 13.49 & 8.82 & 14.14 & 13.49 & 13.25 & 12.22 & 14.12 & 12.99 \\
& SSIM $\uparrow$ & 0.572 & 0.406 & 0.552 & 0.217 & 0.484 & 0.567 & 0.452 & 0.208 & 0.544 & 0.445 \\
& LPIPS $\downarrow$ & 0.717 & 0.692 & 0.612 & 0.671 & 0.696 & 0.725 & 0.688 & 0.891 & 0.646 & 0.704 \\
\hline

\multirow{3}{*}{Luminance-GS~\citep{cui2025luminance}}
& PSNR $\uparrow$ & 11.20 & 7.33 & 14.82 & 8.09 & 8.88 & 9.67 & 11.33 & 9.79 & 9.36 & 10.05 \\
& SSIM $\uparrow$ & 0.465 & 0.362 & 0.536 & 0.411 & 0.372 & 0.532 & 0.462 & 0.293 & 0.463 & 0.433 \\
& LPIPS $\downarrow$ & 0.779 & 0.779 & 0.534 & 0.737 & 0.614 & 0.766 & 0.625 & 0.655 & 0.882 & 0.708 \\
\hline

\multirow{3}{*}{\shortstack{DarkIR-m~\citep{feijoo2025darkir} \\ + 3DGS~\citep{kerbl20233d}}}
& PSNR $\uparrow$ & 15.22 & 12.03 & 18.00 & 13.48 & 12.99 & 14.82 &17.69 & 13.47 & 20.97 &15.41  \\
& SSIM $\uparrow$ & 0.622 & 0.404 & 0.726 & 0.564 & 0.468 & 0.283 &0.608 &\cellcolor{gray!25} 0.299 &\cellcolor{gray!40}\textbf{0.720} &0.522  \\
& LPIPS $\downarrow$ & 0.671 & 0.587 & 0.418 & 0.537 & 0.568 & 0.663 &{0.464} & 0.660 & 0.546 &0.568  \\
\hline

\multirow{3}{*}{\shortstack{Multinex~\citep{brateanu2026multinex} \\ + 3DGS~\citep{kerbl20233d}}}
& PSNR $\uparrow$ & \cellcolor{gray!25}23.06 & \cellcolor{gray!25}19.97 &\cellcolor{gray!25} 19.84 & 17.56 & \cellcolor{gray!25}22.24 & 19.05 &\cellcolor{gray!25}19.53 & 14.02 & 20.43 & \cellcolor{gray!25} 19.52  \\
& SSIM $\uparrow$ & \cellcolor{gray!25}0.736 &\cellcolor{gray!25} 0.604 & 0.711 & \cellcolor{gray!25}0.647 &\cellcolor{gray!40}\textbf{0.736} &\cellcolor{gray!25} 0.701 & 0.624& 0.235 & 0.704& \cellcolor{gray!25}0.633  \\
& LPIPS $\downarrow$ & \cellcolor{gray!25}0.417 &\cellcolor{gray!25}{0.452} & 0.460 & 0.468 &\cellcolor{gray!25} 0.398 &\cellcolor{gray!25} 0.364& \cellcolor{gray!25}0.345 &\cellcolor{gray!25} 0.662 & 0.414 & \cellcolor{gray!25}0.442   \\
\hline

\multirow{3}{*}{\shortstack{DarkIR-m~\citep{feijoo2025darkir} \\ + LITA-GS~\citep{zhou2025lita}}}
& PSNR $\uparrow$ & 16.59 & 14.90 & 15.67 &\cellcolor{gray!40}\textbf{18.85} & 16.43 & \cellcolor{gray!25}19.90 & 17.10 &{14.10} & 21.09 &17.18   \\
& SSIM $\uparrow$ & 0.687 & 0.524 & 0.698 &\cellcolor{gray!40}\textbf{0.671} & 0.697 & 0.698 &\cellcolor{gray!25} 0.667 & 0.287 & 0.713 &0.627  \\
& LPIPS $\downarrow$ & 0.467 & 0.494 & 0.473 & \cellcolor{gray!25}0.466 & 0.439 & 0.370 & 0.375 & 0.531 &\cellcolor{gray!40}\textbf{0.387} &0.445   \\
\hline

\multirow{3}{*}{\shortstack{Multinex~\citep{brateanu2026multinex} \\ + LITA-GS~\citep{zhou2025lita}}}
& PSNR $\uparrow$ & 16.44 & 15.78 & 16.81 & 17.39 & 17.02 & 19.32 & 17.23 & \cellcolor{gray!25}15.32 & 20.59 & 17.32 \\
& SSIM $\uparrow$ & 0.644 & 0.487 &\cellcolor{gray!25} 0.737 & 0.637 & 0.648 & 0.673 & 0.628 & 0.291 & 0.704 & 0.605 \\
& LPIPS $\downarrow$ & 0.505 & 0.502 & \cellcolor{gray!25}0.324 & 0.493 & 0.452 & 0.458 & 0.359 & 0.643 & 0.418 &0.462  \\
\hline

\multirow{3}{*}{\textbf{MLLM-GS (Ours)}}
& PSNR $\uparrow$ & \cellcolor{gray!40}\textbf{24.33} & \cellcolor{gray!40}\textbf{21.39} & 
\cellcolor{gray!40}\textbf{20.31} 
& \cellcolor{gray!25}{18.16} & \cellcolor{gray!40}\textbf{22.94} & \cellcolor{gray!40}\textbf{19.98} & \cellcolor{gray!40}\textbf{20.16} & \cellcolor{gray!40}\textbf{16.21} & \cellcolor{gray!40}\textbf{21.11} & \cellcolor{gray!40}\textbf{20.51}\\
& SSIM $\uparrow$ & \cellcolor{gray!40}\textbf{0.776} & \cellcolor{gray!40}\textbf{0.626} & \cellcolor{gray!40}\textbf{0.740} & \cellcolor{gray!25}{0.649} & \cellcolor{gray!25}{0.673} & \cellcolor{gray!40}\textbf{0.703} & \cellcolor{gray!40}\textbf{0.670} & \cellcolor{gray!40}\textbf{0.294} & \cellcolor{gray!25}0.719 & \cellcolor{gray!40}\textbf{0.650}  \\
& LPIPS $\downarrow$ & \cellcolor{gray!40}\textbf{0.379} & \cellcolor{gray!40}\textbf{0.417} &\cellcolor{gray!40}\textbf{0.306} & \cellcolor{gray!40}\textbf{0.462} & \cellcolor{gray!40}\textbf{0.393} & \cellcolor{gray!40}\textbf{0.359} & \cellcolor{gray!40}\textbf{0.337} & \cellcolor{gray!40}\textbf{0.670} & \cellcolor{gray!25}0.410 & \cellcolor{gray!40}\textbf{0.415} \\
\hline
\end{tabular}
}
\end{table*}

\textbf{User Study:}
\label{User Study}
To further evaluate perceptual quality, we conduct a user study on the synthesized novel views. 
We randomly select ten novel viewpoints and invite $55$ participants to evaluate the corresponding rendered images.
For each viewpoint, the results from different methods are anonymized and presented in random order. 
Participants are asked to select the result with the highest perceived visual quality.
As shown in Fig.~\ref{fig:user_study}(a), our method receives $80.72$\% of the total preferences, substantially outperforming the compared baselines.
This result further supports the advantages of MLLM-GS in synthesizing visually realistic novel views under smoke degradation.

\subsection{Results in Low-Light Scenes}

\textbf{Quantitative Results:}
We compare our method with nine competing methods, including five classic 3D reconstruction methods (3DGS~\citep{kerbl20233d}, LITA-GS~\citep{zhou2025lita}, I2-NeRF~\citep{liu2026i2}, Aleth-NeRF~\citep{cui2024aleth}, and Luminance-GS~\citep{cui2025luminance}), and four restoration-based reconstruction pipelines combining DarkIR-m~\citep{feijoo2025darkir} or Multinex~\citep{brateanu2026multinex} with 3DGS or LITA-GS.
Similar to the smoke scenario, DarkIR-m~\citep{feijoo2025darkir} and Multinex~\citep{brateanu2026multinex} are fine-tuned using the same three paired degraded and clean views per scene.
As shown in Table~\ref{tab:lowlight_exp}, classic 3D reconstruction methods generally achieve limited performance under severe low-light conditions, where insufficient illumination and image noise make accurate geometry and appearance estimation challenging. 
\begin{figure}[t]
\centering
\includegraphics[width=0.9\linewidth]{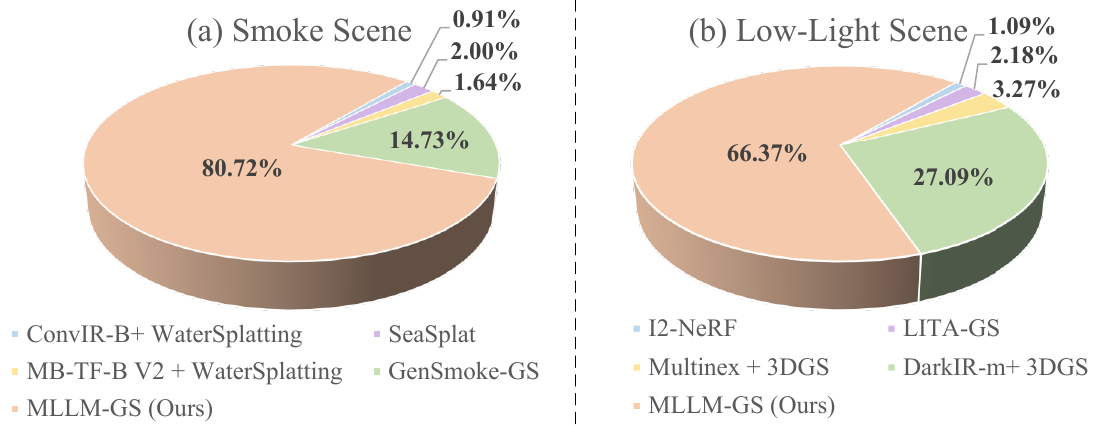}
\caption{User study for smoke and low‑light scenes. }
\label{fig:user_study}
\vspace{-0.4cm}
\end{figure}
Restoration-based reconstruction pipelines improve reconstruction quality by enhancing degraded observations before 3D reconstruction, achieving average PSNR values ranging from $15.41$ dB to $19.52$ dB.
However, residual noise, color distortions, and inconsistencies in the enhanced views may propagate into subsequent reconstruction, limiting the final rendering quality. 
In contrast, our method achieves the best average PSNR, SSIM, and LPIPS.
Compared with the strongest baseline, our method improves PSNR and SSIM by $0.99$ dB and $0.017$, respectively, while reducing LPIPS by $0.027$.
These results demonstrate that MLLM-GS achieves superior reconstruction quality under severe low-light conditions, particularly in terms of pixel-level fidelity.

\textbf{Qualitative Results:}
Fig.~\ref{fig:vis_lowlight_compare} illustrates qualitative comparisons of NVS results in representative low-light scenes. 
Standard 3DGS~\citep{kerbl20233d} produces severely underexposed renderings with limited visible structures and textures.
LITA-GS~\citep{zhou2025lita} improves overall brightness but still exhibits noticeable noise and color distortions, particularly around object boundaries and textured regions. 
Restoration-based reconstruction pipelines combining DarkIR-m~\citep{feijoo2025darkir} or Multinex~\citep{brateanu2026multinex} with 3DGS further improve scene visibility, although uneven illumination, color inconsistencies, and loss of fine-grained details remain visible in some regions. 
In contrast, our method produces visually faithful renderings with clearer object structures, more consistent colors, and better-preserved local details. 
The reconstructed results more closely resemble the ground-truth images in the examples shown, particularly around object boundaries and textured regions. 
These observations demonstrate the effectiveness of MLLM-GS in recovering scene appearance and preserving visual details under severe low-light degradation.
\begin{figure*}[!t]
\centering
\includegraphics[width=1\textwidth]{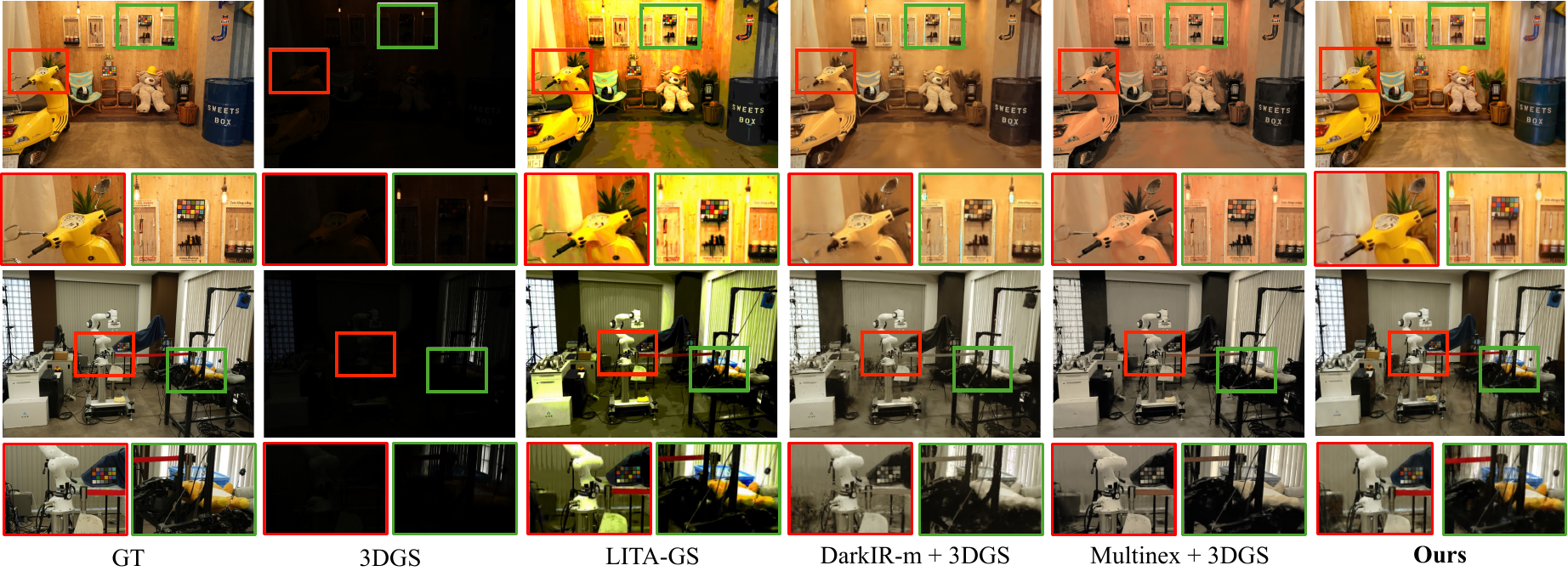}
\vspace{-0.5cm}
\caption{Qualitative comparison of NVS results under low-light scenes.}
\label{fig:vis_lowlight_compare}
\vspace{-0.25cm}
\end{figure*}

\textbf{User Study:}
To further evaluate perceptual quality, we conduct a user study in low-light scenes following the evaluation protocol described in Section~\ref{User Study}. 
We randomly select ten novel views and invite $55$ participants to assess the rendered results. 
As shown in Fig.~\ref{fig:user_study}(b), our method receives $66.37$\% of the total preferences, substantially outperforming the compared methods. 
These results indicate that the proposed MLLM-GS produces perceptually realistic novel views under severe low-light conditions.

\begin{table}[t]
\centering
\vspace{-0.2cm}
\footnotesize
\caption{Efficiency comparison of 3D reconstruction methods. The training steps for all methods are 30,000. ``N/A'' indicates that the corresponding metric is not applicable to the method.}
\setlength{\tabcolsep}{6pt}
\begin{tabular}{|l|ccc|}
\hline\thickhline
Method & Training $\downarrow$ & \#Gs $\downarrow$ & VRAM $\downarrow$ \\ \hline
\multicolumn{4}{|l|}{\textit{Akikaze (Smoke)}} \\
\hline
3DGS & 14m44s & 2.26M & 8.15GiB\\ 
SeaSplat & 71m59s & 0.63M & 7.06GiB \\ 
WaterSplatting & 19m30s & \cellcolor{gray!40}\textbf{0.03M} & \cellcolor{gray!25}5.06GiB\\ 
I2-NeRF & 64m05s & N/A & 5.13GiB \\
SeaThruNeRF &90m59s & N/A & 30.77GiB \\
Faster-GS  &  \cellcolor{gray!40}\textbf{8m37s} & \cellcolor{gray!25}0.06M  & 6.36GiB \\ \hline
\textbf{MLLM-GS (Ours)} & \cellcolor{gray!25}10m06s & 0.46M & \cellcolor{gray!40}\textbf{1.98GiB} \\
\hline
\multicolumn{4}{|l|}{\textit{BlueHawaii (Low-Light)}} \\ \hline
3DGS & 15m35s & 3.29M & 9.64GiB \\
LITA-GS & 234m53s & 0.32M & 36.11GiB \\
I2-NeRF & 121m27s & N/A & 26.59GiB \\ 
Aleth-NeRF & 39m35s & N/A & 18.69GiB \\
Luminance-GS & 88m32s & 1.31M & \cellcolor{gray!25} 3.07GiB \\
Faster-GS  & \cellcolor{gray!40}\textbf{9m19s}  & \cellcolor{gray!40}\textbf{0.05M}  & 6.79GiB \\ \hline
\textbf{MLLM-GS (Ours)} & \cellcolor{gray!25}9m38s & \cellcolor{gray!25}0.17M & \cellcolor{gray!40}\textbf{2.40GiB} \\
\hline
\end{tabular}
\vspace{-0.4cm}
\label{tab:efficiency}
\end{table}

\subsection{Efficiency Analysis}
In this section, we evaluate the reconstruction efficiency of MLLM-GS and competing methods on smoke and low-light scenes in terms of training time, Gaussian count, and GPU memory consumption. 
As shown in Table~\ref{tab:efficiency}, MLLM-GS achieves high training efficiency under both degradation conditions.
On the smoke scene, MLLM-GS completes $30$K training iterations in $10$m$06$s, achieving a $1.46\times$ speedup over 3DGS~\citep{kerbl20233d} and a $9.0\times$ speedup over SeaThruNeRF~\citep{levy2023seathru}. 
On the low-light scene, it requires $9$m$38$s, achieving a $1.62\times$ speedup over 3DGS and a $24.4\times$ speedup over LITA-GS~\citep{zhou2025lita}. 
Although Faster-GS~\citep{hahlbohm2026faster} achieves slightly shorter training times, MLLM-GS provides substantially better reconstruction quality while maintaining comparable optimization efficiency. 
Moreover, compared with standard 3DGS, MLLM-GS reduces the Gaussian count by approximately $80$\% and $95$\% on the smoke and low-light scenes, respectively.
It also achieves the lowest GPU memory consumption among all compared methods, requiring only $1.98$ GiB and $2.40$ GiB on the two scenes, respectively.

\begin{figure}[t!]
\vspace{-0.2cm}
\centering
\includegraphics[width=1\linewidth]{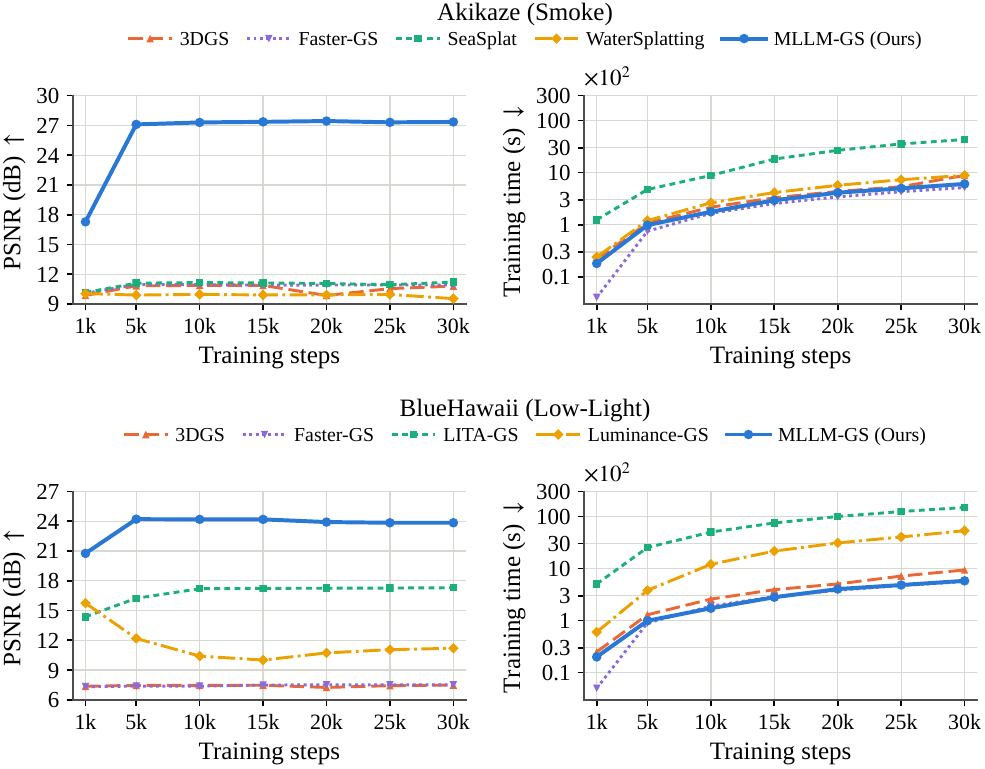}
\caption{Efficiency analysis on a smoke scene (Akikaze) and a low-light scene (BlueHawaii). }
\label{fig:efficiency_curve}
\vspace{-0.3cm}
\end{figure}

Fig.~\ref{fig:efficiency_curve} further compares the convergence behavior and computational efficiency of different methods. 
After only $5$K training iterations, MLLM-GS achieves approximately $27$ dB PSNR on Akikaze and $24$ dB on BlueHawaii, requiring around $100$ s. 
These results surpass those achieved by all compared methods after $30$K iterations, demonstrating that MLLM-GS achieves higher reconstruction quality with substantially fewer iterations. 
In contrast, Faster-GS~\citep{hahlbohm2026faster} achieves only approximately $11$ dB and $7$ dB PSNR on the two scenes, respectively, even after $30$K iterations. 
This indicates that accelerating Gaussian optimization alone cannot effectively address the reconstruction challenges caused by severe image degradations.
By combining MLLM prior-guided view enhancement with efficient Gaussian optimization, MLLM-GS achieves a favorable balance between reconstruction quality and computational efficiency.

\begin{table}[t]
\footnotesize
\centering
\caption{Ablation of different MLLM priors on Akikaze under smoke degradation.}
\tabcolsep 3.5pt
\begin{tabular}{|l|ccc|}
\hline\thickhline
Method & PSNR $\uparrow$ & SSIM $\uparrow$ & LPIPS $\downarrow$ \\
\hline
Wan2.7-Image & 22.13 & 0.714 & 0.381 \\
Qwen-Image-3.0 & 22.29 & 0.713 & 0.431 \\
Seedream-5.0-Lite & 22.21 & 0.675 & 0.421 \\ 
GPT-Image-2 & 19.06 & 0.644  & 0.482 \\
Qwen-Image-Edit& 18.09 & 0.621 & 0.630 \\
\hline
\cellcolor{gray!40}\textbf{Nano Banana Pro (Ours)} & \cellcolor{gray!40}\textbf{27.44} & \cellcolor{gray!40}\textbf{0.761} & \cellcolor{gray!40}\textbf{0.234} \\
\hline
\end{tabular}
\label{tab:mllm_prior_smoke}
\vspace{-0.35cm}
\end{table}
\begin{figure}[t]
\centering
\vspace{-0.5cm}
\includegraphics[width=1\linewidth]{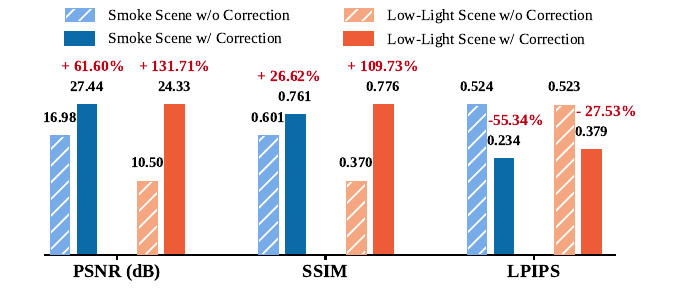}
\vspace{-0.55cm}
\caption{Ablation study on the correction network. }
\label{fig:corr_vis}
\vspace{-0.4cm}
\end{figure}

\subsection{Ablation Studies}

\textbf{Effectiveness of MLLM Prior:}
To evaluate the effectiveness of MLLM prior-guided view enhancement, we compare our framework with standard 3DGS~\citep{kerbl20233d} and two-stage restoration-based reconstruction methods. 
Standard 3DGS~\citep{kerbl20233d} directly optimizes the scene representation from degraded observations without image enhancement. 
As shown in Tables~\ref{tab:smoke_exp} and \ref{tab:lowlight_exp}, its average PSNR is below $12$ dB on both smoke and low-light scenes, indicating its limited ability to reconstruct clean views from severely degraded images. 
In contrast, MLLM-GS improves the average PSNR by more than $10$ dB on both scene types and achieves substantial improvements in SSIM and LPIPS. 
Furthermore, although two-stage methods improve reconstruction quality by restoring degraded images before 3D reconstruction, their performance remains inferior to that of MLLM-GS. 
This is mainly because degradation-specific restoration models may struggle to recover reliable structural and appearance information under complex imaging conditions.
In contrast, our framework leverages the general visual priors of a pretrained MLLM to enhance degraded observations across different degradation types. 
The correction network further reduces residual spatial misalignment and color deviations, providing more reliable pseudo-GT supervision for subsequent 3D reconstruction. 
These results demonstrate the effectiveness of MLLM prior-guided view enhancement in improving NVS quality under severe degradations.

\begin{table}[t]
\centering
\footnotesize
\caption{Ablation of the proposed components on Akikaze under smoke degradation.
}
\tabcolsep 2.8pt
\begin{tabular}{|l|ccc|}
\hline\thickhline
Method  & Training $\downarrow$ & \#Gs$\downarrow$ & VRAM $\downarrow$ \\
\hline
3DGS  & 14m44s & 2.26M & 8.15GiB \\
+ Rasterization  & 11m27s & 0.49M & 2.57GiB \\
\rowcolor{gray!25} \quad + Memory Layout  & 10m42s & 0.47M & 2.12GiB \\
\quad\quad \cellcolor{gray!40}{+ Parameter Updates}  & \cellcolor{gray!40}\textbf{10m06s} & \cellcolor{gray!40}\textbf{0.46M} & \cellcolor{gray!40}\textbf{1.98GiB} \\
\hline
\end{tabular}
\label{tab:ablation_components_akikaze}
\vspace{-0.5cm}
\end{table}

\begin{figure}[t]
\centering
\vspace{-0.5cm}
\includegraphics[width=0.92\linewidth]{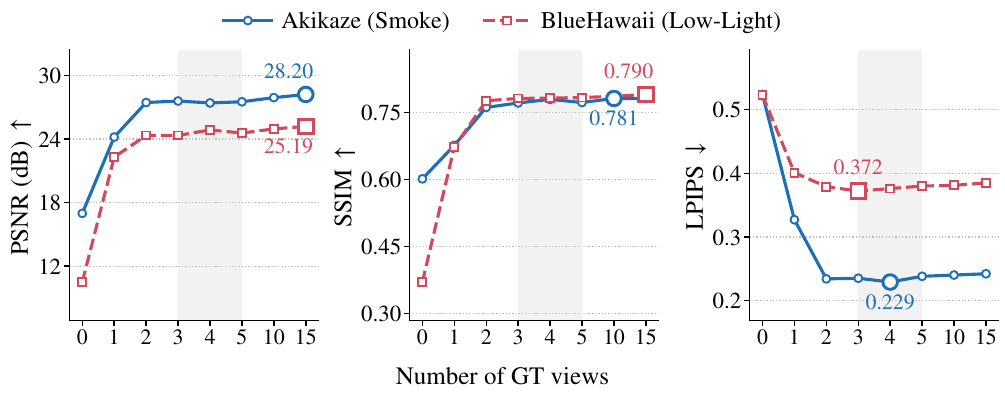}
\vspace{-0.2cm}
\caption{Ablation on the number of GT views.
%{\color{red}update the value, 27.44 / 0.761}
}
\label{fig:view_ablation}
\vspace{-0.5cm}
\end{figure}

\begin{figure*}[!t]
\centering
\includegraphics[width=1\textwidth]{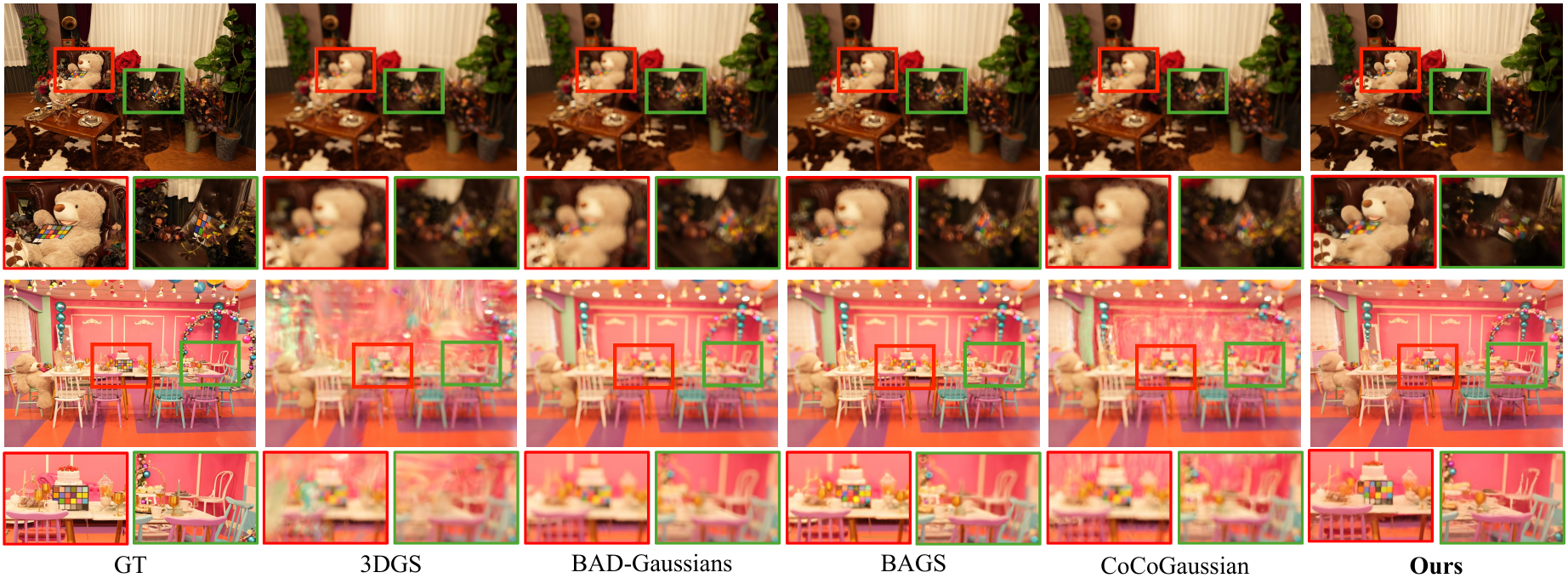}
\caption{Qualitative comparison of NVS results under strong defocus blur scenes.}
\label{fig:strong_defocus_compare}
\vspace{-0.5cm}
\end{figure*}

\textbf{Compatibility of MLLM Priors:}
This section evaluates the compatibility of our framework with different MLLM priors. 
Specifically, we replace the default Nano Banana Pro~\citep{raisinghani2025nano} with five alternative models, including Wan2.7-Image~\citep{wan2.7image}, Qwen-Image-3.0~\citep{qwen3.0image}, Seedream-5.0-Lite ~\citep{seed}, GPT-Image-2~\citep{openai2026c2pa}, and Qwen-Image-Edit~\citep{wu2025qwenimage}, and evaluate their performance on the Akikaze smoke scene.
As shown in Table~\ref{tab:mllm_prior_smoke}, all five alternatives can be integrated into our framework, achieving PSNR values ranging from $18.09$ dB to $22.29$ dB.
However, their reconstruction performance varies, indicating that the quality of the underlying generative prior influences the final NVS results. 
The default model achieves the best performance across all three metrics, with a PSNR of $27.44$ dB, an SSIM of $0.761$, and an LPIPS of $0.234$. 
These results demonstrate that our framework can accommodate different generative priors, while stronger priors can further improve reconstruction quality.

\textbf{Necessity of the Correction:}
To evaluate the necessity of the correction network, we compare the full framework with a variant that directly uses MLLM-enhanced images for 3D reconstruction without correction.
Fig.~\ref{fig:corr_vis} presents the quantitative results on the Akikaze (smoke) and BlueHawaii (low-light) scenes.
Introducing the correction network consistently improves all three metrics under both degradation conditions. 
On the Akikaze scene, PSNR increases from $16.98$ to $27.44$ dB, SSIM improves from $0.601$ to $0.761$, and LPIPS decreases from $0.524$ to $0.234$. 
The improvements are more pronounced on the BlueHawaii scene, where PSNR increases from $10.50$ to $24.33$ dB, SSIM improves from $0.370$ to $0.776$, and LPIPS decreases from $0.523$ to $0.379$. 
These improvements indicate that MLLM-enhanced images still contain residual spatial misalignment and color deviations that can affect subsequent 3D reconstruction.
By reducing these errors, the correction network provides more accurate and consistent pseudo-GT supervision, leading to improved reconstruction quality.

\begin{figure*}[!t]
\centering
\includegraphics[width=1\textwidth]{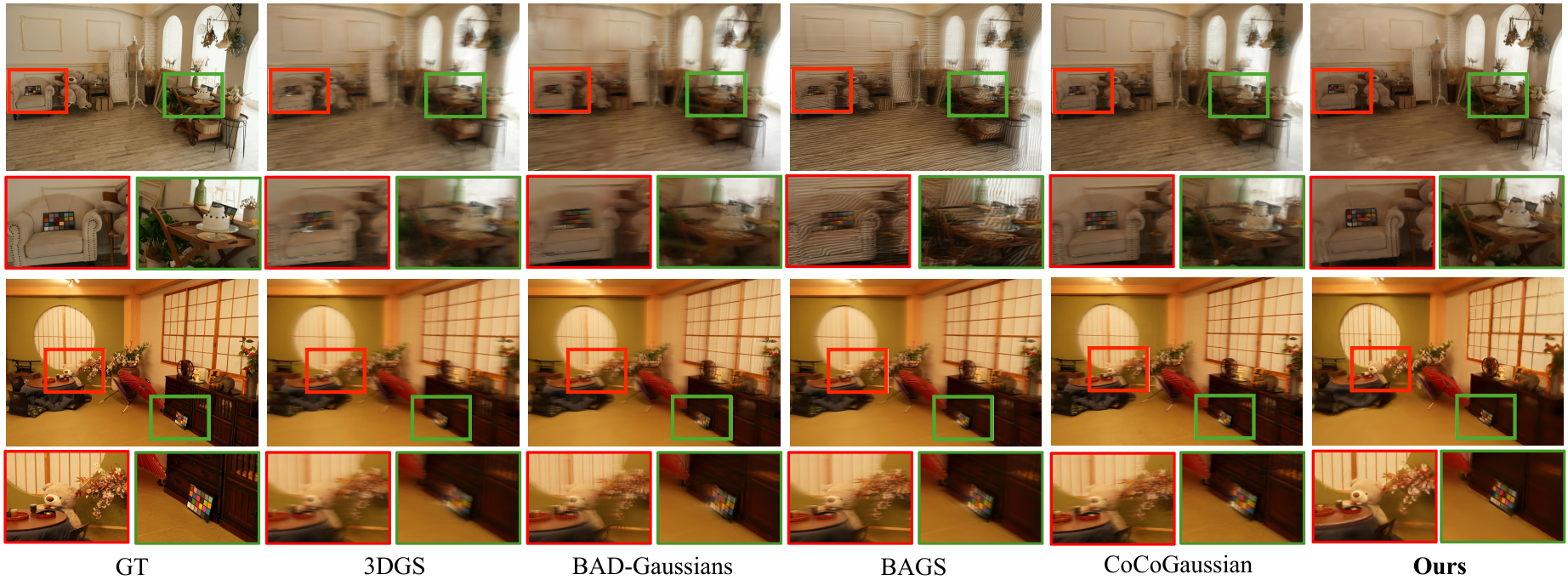}
\caption{Qualitative comparison of NVS results under strong motion blur scenes.
}
\label{fig:strong_motion_compare}
\vspace{-0.4cm}
\end{figure*}

\textbf{Impact of the GT Views Number:}
This section investigates the impact of the number of paired clean GT views on reconstruction quality under smoke and low-light conditions. 
As shown in Fig.~\ref{fig:view_ablation}, introducing a small number of GT views substantially improves reconstruction performance, with the largest gains occurring between zero and three views. 
Specifically, increasing the number of GT views from zero to three improves PSNR by more than $13$ dB on the low-light scene and approximately $10$ dB on the smoke scene.
Similar improvements are observed in SSIM and LPIPS. 
As more GT views are introduced, PSNR and SSIM continue to improve, but the gains become smaller. 
For example, the additional PSNR improvement beyond five GT views is less than $0.6$ dB on the low-light scene.
Meanwhile, LPIPS achieves its best performance with approximately four to five GT views and does not consistently improve as more views are added. 
These results indicate that a few paired clean views already provide effective supervision for the correction network. 
Increasing the number of GT views yields diminishing improvements in PSNR and SSIM, while perceptual quality does not necessarily benefit from additional supervision.

\textbf{Effectiveness of Rasterization Optimization:}
To evaluate the effectiveness of rasterization optimization, we replace the standard 3DGS~\citep{kerbl20233d} rasterization pipeline while keeping all other settings unchanged.
As shown in Table~\ref{tab:ablation_components_akikaze}, the number of Gaussians decreases from $2.26$M to $0.49$M, GPU memory consumption drops from $8.15$ GiB to $2.57$ GiB, and training time is reduced by $22.3$\%.
The opacity-aware rectangular bounding boxes reduce unnecessary Gaussian–tile intersections by excluding regions with negligible contributions, thereby lowering sorting and forward rasterization costs. 
Meanwhile, the bucket-based backward propagation strategy reconstructs the blending process using precomputed boundary states, reducing redundant gradient computations and memory accesses.
In this way, these designs improve rasterization efficiency and reduce the computational overhead of 3D reconstruction.

\textbf{Effectiveness of Memory Layout Optimization:}
To evaluate the effectiveness of memory layout optimization, we introduce Morton-based Gaussian reordering on top of the rasterization optimization.
As shown in Table~\ref{tab:ablation_components_akikaze}, the number of Gaussians, memory footprint, and training time are all reduced, with reductions ranging from $4$\% to $18$\%.
These improvements demonstrate the benefits of optimizing Gaussian memory organization. 
During adaptive densification, newly generated Gaussians are appended to parameter buffers, causing spatially neighboring Gaussians to become scattered across memory. 
Morton reordering organizes Gaussians according to their spatial positions, improving memory-access locality and reducing redundant memory accesses during optimization.
By periodically reorganizing Gaussian parameters, this strategy further improves efficiency with limited reordering overhead.

\textbf{Effectiveness of Parameter Update Optimization:}
We further evaluate the effectiveness of parameter update optimization on top of the rasterization and memory layout optimizations. 
As shown in Table~\ref{tab:ablation_components_akikaze}, training time decreases from $10m42$s to $10m06$s, while GPU memory consumption is reduced from $2.12$ GiB to $1.98$ GiB.
The number of Gaussians also decreases slightly from $0.47$M to $0.46$M. 
The training efficiency improvement is mainly attributed to the fused parameter update strategy, which combines the updates of different Gaussian parameter groups into a single CUDA kernel. 
This reduces kernel-launch overhead and redundant global-memory accesses while preserving the underlying Adam update rules. 
Overall, combining rasterization, memory layout, and parameter update optimizations reduces the number of Gaussians by approximately $80$\%, GPU memory consumption by $76$\%, and training time by $31$\% compared with standard 3DGS~\citep{kerbl20233d}, demonstrating the effectiveness of the proposed fast reconstruction branch.

\begin{table}[t]
\footnotesize
\vspace{-0.25cm}
\centering
\caption{Effectiveness of fast 3D scene reconstruction branch.}
\tabcolsep 7pt
\begin{tabular}{|l|ccc|}
\hline\thickhline
Method & PSNR $\uparrow$ & SSIM $\uparrow$ & LPIPS $\downarrow$ \\
\hline
\multicolumn{4}{|l|}{\textit{Akikaze (Smoke)}} \\
\hline
+Faster-GS& 24.85 & 0.710  & 0.340 \\
\hline
\cellcolor{gray!40}\textbf{MLLM-GS (Ours)} & \cellcolor{gray!40}\textbf{27.44} & \cellcolor{gray!40}\textbf{0.761} & \cellcolor{gray!40}\textbf{0.234} \\
\hline
\multicolumn{4}{|l|}{\textit{BlueHawaii (Low-Light)}} \\ \hline
+Faster-GS& 20.03 & 0.643 &  0.397\\
\hline
\cellcolor{gray!40}\textbf{MLLM-GS (Ours)} & \cellcolor{gray!40}\textbf{24.33} & \cellcolor{gray!40}\textbf{0.776} & \cellcolor{gray!40}\textbf{0.379} \\
\hline
\end{tabular}
\label{tab:3d_branch_smoke}
\vspace{-0.65cm}
\end{table}

\begin{figure*}[!t]
\centering
\vspace{-0.4cm}
\includegraphics[width=1\textwidth]{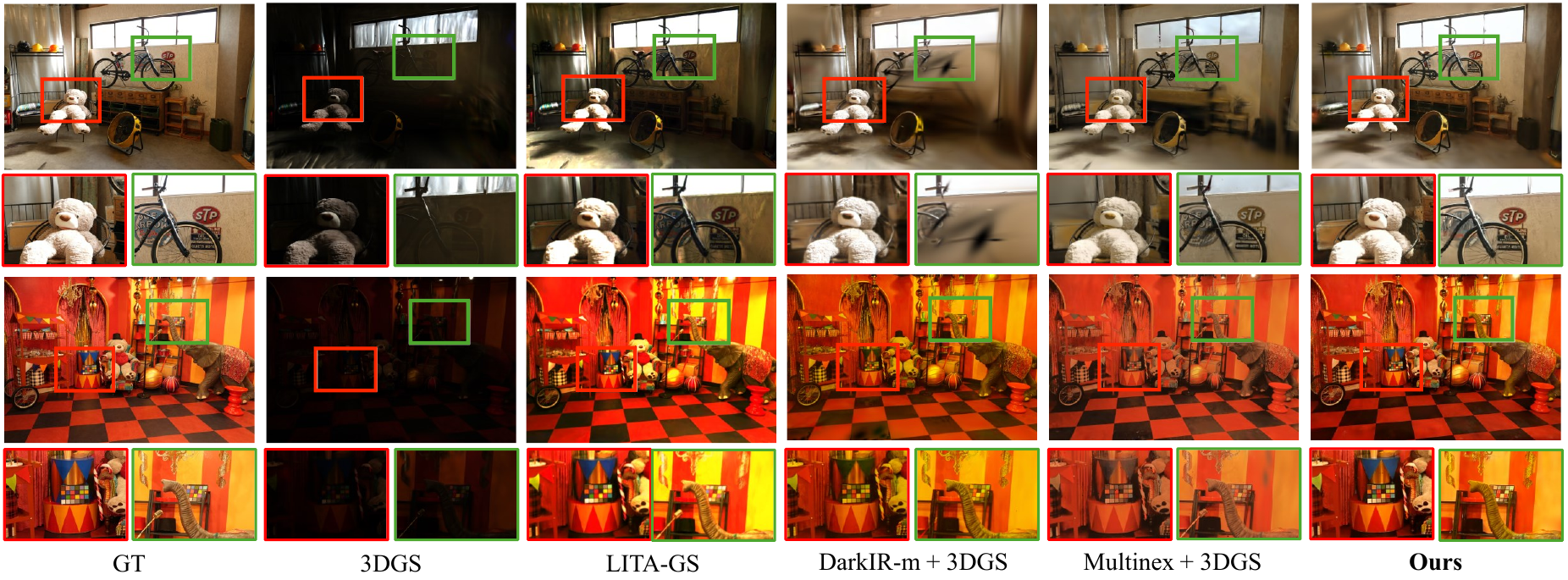}
\caption{Qualitative comparison of NVS results under varying exposure scenes.
}
\label{fig:varyexp_compare}
\vspace{-0.45cm}
\end{figure*}
\textbf{Effectiveness of fast 3D scene reconstruction branch:}
To evaluate the effectiveness of the proposed fast 3D reconstruction branch, we compare MLLM-GS with a variant that replaces this branch with Faster-GS~\citep{hahlbohm2026faster} while keeping the view enhancement branch unchanged. 
Experiments are conducted on the Akikaze (Smoke) and BlueHawaii (Low-Light) scenes. 
As shown in Table~\ref{tab:3d_branch_smoke}, MLLM-GS consistently outperforms the Faster-GS variant across all three metrics in both scenes.
These results demonstrate the effectiveness of the proposed fast reconstruction branch in supporting high-quality 3D scene reconstruction under challenging environmental conditions.

\subsection{Experimental Results for More Degradation Scenarios}

To further evaluate the generalization capability of MLLM-GS across diverse degradation conditions, we conduct additional experiments under strong defocus blur, strong motion blur, and varying exposure.
These scenarios introduce different types of structural and appearance distortions, providing a more comprehensive evaluation of the proposed framework.

\textbf{Results of Strong Defocus Blur:}
As shown in Fig.~\ref{fig:strong_defocus_compare}, 3DGS~\citep{kerbl20233d} produces severely degraded renderings with noticeable color shifts, banding artifacts, and blurred scene structures. 
BAD-Gaussians~\citep{zhao2024bad}, BAGS~\citep{peng2024bags}, and CoCoGaussian~\citep{lee2025cocogaussian} recover the overall scene structure, but their results still exhibit blurred object boundaries and missing fine details.
For example, adjacent patches on the color chart become difficult to distinguish, while the fine textures of the toy are largely lost. 
In contrast, our method produces clearer scene structures and better-preserved textures. The color chart retains well-defined boundaries, and the toy exhibits more recognizable fur textures and facial details. 
These results demonstrate that MLLM-GS can effectively reconstruct scene structures and fine details from observations affected by strong defocus blur.

\textbf{Results of Strong Motion Blur:}
Fig.~\ref{fig:strong_motion_compare} presents qualitative comparisons of NVS results under strong motion blur. 
3DGS~\citep{kerbl20233d} produces severe motion-related artifacts, resulting in distorted scene structures and blurred object boundaries. 
BAD-Gaussians~\citep{zhao2024bad} and CoCoGaussian~\citep{lee2025cocogaussian} recover the overall scene layout, but their results still suffer from blurred textures and missing fine details. 
BAGS~\citep{peng2024bags} also recovers the scene structure but introduces noticeable stripe artifacts. 
In contrast, our method generates clearer novel views with more accurate object boundaries and better-preserved textures. 
In particular, the color chart retains distinguishable patches, while the fur textures of the toy and the structures of small objects remain clearly recognizable. 
These results demonstrate the effectiveness of MLLM-GS in preserving scene structures and fine details under strong motion blur.

\textbf{Results of Varying Exposure:}
Fig.~\ref{fig:varyexp_compare} presents qualitative comparisons of NVS results under varying exposure conditions. 
3DGS~\citep{kerbl20233d} is sensitive to exposure variations, producing severely underexposed renderings with substantial detail loss. 
LITA-GS~\citep{zhou2025lita} improves the overall brightness, but noticeable color and contrast deviations remain in some regions.
DarkIR-m~\citep{feijoo2025darkir} + 3DGS and Multinex~\citep{brateanu2026multinex} + 3DGS further improve scene visibility, but their results still exhibit blurred textures and local color distortions.
In contrast, our method produces novel views with more natural brightness, improved color fidelity, and clearer scene details.
In particular, the structures and textures of objects such as the teddy bear, bicycle, and indoor furniture are better preserved. 
These results demonstrate that MLLM-GS can effectively handle exposure variations and generate high-quality novel views with more faithful scene appearance.

\section{Conclusion}
This paper proposes MLLM-GS, an MLLM prior-guided framework for fast novel view synthesis in extreme environments. 
It consists of two branches: MLLM prior-guided view enhancement and fast 3D scene reconstruction. 
The former combines a frozen MLLM with a lightweight correction network to reduce pixel misalignment and color distortions, providing reliable pseudo-GT supervision using only a few paired clean views. 
The latter accelerates 3DGS optimization from three aspects: rasterization, memory layout, and parameter updates.
Extensive experiments demonstrate that MLLM-GS achieves high-quality novel view synthesis across diverse severe degradations while substantially reducing reconstruction time and GPU memory consumption.

\backmatter

\section*{Declarations}

\begin{itemize}

\item \textbf{Funding:} This work was supported by the Yangtze River Delta Science and Technology Innovation Community Joint Research (Basic Research) Project (2025CSJZN01600) and the New Generation of Artificial Intelligence National Science and Technology Major Project (2025ZD0123303). 

\item \textbf{Data availability:} %Data sharing is not applicable to this article as no new data were created or analyzed in this study.
The datasets analyzed in this study are publicly available. The RealX3D dataset is available at \url{https://huggingface.co/datasets/ToferFish/RealX3D}. 
\item \textbf{Competing interests:} The authors declare no competing interests.
\end{itemize}

\nocite{*}
\balance
\bibliography{sn-bibliography}% common bib file

\end{document}